\documentclass[journal]{IEEEtran}
\usepackage{amsfonts}
\usepackage{multirow}
\usepackage{subfigure}
\usepackage{cite}

\usepackage{cite}
\ifCLASSINFOpdf
  \usepackage[pdftex]{graphicx}
  \DeclareGraphicsExtensions{.pdf,.jpeg,.png}
\else
\fi
\usepackage{amssymb,amsmath,bm}

\begin{document}
%
% paper title
% Titles are generally capitalized except for words such as a, an, and, as,
% at, but, by, for, in, nor, of, on, or, the, to and up, which are usually
% not capitalized unless they are the first or last word of the title.
% Linebreaks \\ can be used within to get better formatting as desired.
% Do not put math or special symbols in the title.
\title{A Sequential Neural Encoder with Latent Structured Description for Modeling Sentences}
%
%
% author names and IEEE memberships
% note positions of commas and nonbreaking spaces ( ~ ) LaTeX will not break
% a structure at a ~ so this keeps an author's name from being broken across
% two lines.
% use \thanks{} to gain access to the first footnote area
% a separate \thanks must be used for each paragraph as LaTeX2e's \thanks
% was not built to handle multiple paragraphs
%

\author{Yu-Ping~Ruan,~Qian~Chen,~and~Zhen-Hua~Ling,~\IEEEmembership{Member,~IEEE}% <-this % stops a space
\thanks{This work was supported in part by the Science and Technology Development of Anhui Province, China (Grant No. 2014z02006), the Fundamental Research Funds for the Central Universities (Grant
No. WK2350000001), the Strategic Priority Research Program of the Chinese Academy of Sciences (Grant No. XDB02070006),
and the National Natural Science Foundation of China (Grant No. U1636201).}
\thanks{Y.-P. Ruan, Q. Chen, and Z.-H. Ling are with the National Engineering Laboratory of Speech and Language Information Processing,
University of Science and Technology of China, Hefei, 230027, China (e-mail: ypruan@mail.ustc.edu.cn, cq1231@mail.ustc.edu.cn, zhling@ustc.edu.cn).}}

% note the % following the last \IEEEmembership and also \thanks -
% these prevent an unwanted space from occurring between the last author name
% and the end of the author line. i.e., if you had this:
%
% \author{....lastname \thanks{...} \thanks{...} }
%                     ^------------^------------^----Do not want these spaces!
%
% a space would be appended to the last name and could cause every name on that
% line to be shifted left slightly. This is one of those "LaTeX things". For
% instance, "\textbf{A} \textbf{B}" will typeset as "A B" not "AB". To get
% "AB" then you have to do: "\textbf{A}\textbf{B}"
% \thanks is no different in this regard, so shield the last } of each \thanks
% that ends a line with a % and do not let a space in before the next \thanks.
% Spaces after \IEEEmembership other than the last one are OK (and needed) as
% you are supposed to have spaces between the names. For what it is worth,
% this is a minor point as most people would not even notice if the said evil
% space somehow managed to creep in.

% The paper headers
\markboth{Journal of \LaTeX\ Class Files,~Vol.~14, No.~8, August~2015}%
{Shell \MakeLowercase{\textit{et al.}}: Bare Demo of IEEEtran.cls for IEEE Journals}
% The only time the second header will appear is for the odd numbered pages
% after the title page when using the twoside option.
%
% *** Note that you probably will NOT want to include the author's ***
% *** name in the headers of peer review papers.                   ***
% You can use \ifCLASSOPTIONpeerreview for conditional compilation here if
% you desire.

% If you want to put a publisher's ID mark on the page you can do it like
% this:
%\IEEEpubid{0000--0000/00\$00.00~\copyright~2015 IEEE}
% Remember, if you use this you must call \IEEEpubidadjcol in the second
% column for its text to clear the IEEEpubid mark.

% use for special paper notices
%\IEEEspecialpapernotice{(Invited Paper)}

% make the title area
\maketitle

% As a general rule, do not put math, special symbols or citations
% in the abstract or keywords.
\begin{abstract}
In this paper, we propose a sequential neural encoder with latent structured description (SNELSD) for modeling sentences.
This model introduces latent chunk-level representations into conventional sequential neural encoders, i.e., recurrent neural networks (RNNs) with long short-term memory (LSTM) units,
to consider the compositionality of languages in semantic modeling.
An SNELSD model has a hierarchical structure that includes a detection layer and a description layer.
The detection layer predicts the  boundaries of latent word chunks in an input sentence and derives a chunk-level vector for each word.
The description layer utilizes modified LSTM units to process these chunk-level vectors in a recurrent manner and produces sequential encoding outputs.
These output vectors  are further concatenated with word vectors or the outputs of a chain LSTM encoder to obtain the final sentence representation.
%which is inspired by human cognition mechanism that people tend to split sentences into several essential target-related parts while reading.
%Unlike Tree-LSTM utilizing the syntactic parsing structure information, the SNELS model can learn to 'chunk' sentences into word blocks
All the model parameters are learned in an end-to-end manner without a dependency on additional text chunking or syntax parsing.
%In other words, it can exploit the latent structure information in sentences which is task-dependent. Further more, the SNELS has a 2-layer hierarchical chain structure, which means that it can own the comparative efficiency in computation to LSTMs models but also can utilize some structure information that flat-chain structure's LSTMs models cannot.
A natural language inference (NLI) task and a sentiment analysis (SA) task are adopted to evaluate the performance of our proposed model.
The experimental results demonstrate the effectiveness of the proposed SNELSD model on exploring task-dependent chunking patterns during the semantic modeling of sentences.
Furthermore, the proposed method achieves better performance than conventional chain LSTMs and tree-structured LSTMs on both tasks.

%In practical test on natural language inference (NLI) and sentiment analysis (SA) tasks, the SNELS model can truly capture some useful and regular 'chunking' pattern that consistent to human cognition process on corresponding task in some way. Moreover, the 'chunking' pattern of SNELS on these two tasks are apparently different, which means that our SNELS can utilize the task-dependent latent structure information. Anyway, SNELS can outperform the ordinary LSTMs and Tree-LSTM models on NLI and do not show weak performance than those models on SA task. On the whole, the SNELS model is promising to be fitted to natural language processing task.
\end{abstract}

% Note that keywords are not normally used for peerreview papers.
\begin{IEEEkeywords}
recurrent neural network, long short-term memory, sentence modeling, syntax structure.
\end{IEEEkeywords}

% For peer review papers, you can put extra information on the cover
% page as needed:
% \ifCLASSOPTIONpeerreview
% \begin{center} \bfseries EDICS Category: 3-BBND \end{center}
% \fi
%
% For peerreview papers, this IEEEtran command inserts a page break and
% creates the second title. It will be ignored for other modes.
\IEEEpeerreviewmaketitle

\section{Introduction}\label{intro}
% The very first letter is a 2 line initial drop letter followed
% by the rest of the first word in caps.
%
% form to use if the first word consists of a single letter:
% \IEEEPARstart{A}{demo} file is ....
%
% form to use if you need the single drop letter followed by
% normal text (unknown if ever used by the IEEE):
% \IEEEPARstart{A}{}demo file is ....
%
% Some journals put the first two words in caps:
% \IEEEPARstart{T}{his demo} file is ....
%
% Here we have the typical use of a "T" for an initial drop letter
% and "HIS" in caps to complete the first word.
\IEEEPARstart{S}{ince} sentence modeling serves as the basis for a wide range of natural language processing (NLP) tasks,
many sentence encoders have been developed to produce vector representations for describing sentence meanings.
Traditionally, the representation of a sentence can be derived based on simple statistics and linguistic rules (e.g., bag-of-words or bag-of-n-grams \cite{harris1954distributional}).
However, these models suffer from the lack of context and word order information.
With the development of word embedding and deep learning techniques,
the focus on sentence modeling has shifted to deriving the compositional sentence representation from a sequence of word vectors using neural networks \cite{sutskever2014sequence},
such as convolutional neural networks (CNNs) \cite{kim2014convolutional,kalchbrenner2014convolutional} and recurrent neural networks (RNNs).
% and tree-structure recursive neural network (TreeRNNs).

%CNNs process text using a group of fixed-size filters sliding along the sentences.
%This kind of networks have been popularly applied to text classification tasks \cite{kim2014convolutional,kalchbrenner2014convolutional}, such as topic classification and sentiment analysis.
%However, CNNs suffer from the deficiency of missing word order information of within a filter window.
%In current sentence encoding models, the most effective and frequently used model is RNNs,
RNNs are currently the most popular sentence encoding models, and they process sentences word by word using a chain structure.
%Because of their capability to retain the history information from an arbitrarily long context window\cite{lipton2015} and  their high efficiency in model training,
Ideally, RNNs own the capability to retain the history information from an arbitrarily long context window\cite{lipton2015}. However, the problem of vanishing and exploding gradients during model training make it difficult for RNNs to learn long-range dependencies\cite{DBLP:journals/tnn/BengioSF94,hochreiter2001gradient}.
To overcome the deficiency of RNNs, the ones that use long short-term memory (LSTM) units \cite{hochreiter1997long} was proposed and have been successfully applied to many NLP tasks,
including dependency parsing \cite{Kiperwasser2016Simple}, named entity recognition \cite{lample2016neural,huang2015bidirectional}, question answering \cite{andreas2016deep,ren2015exploring}, machine translation \cite{bahdanau2014neural,luong2015effective,Ling2015Character}, and so forth.
However, due to their flat chain structure, LSTM-RNNs are incapable of utilizing syntactic information, which is intrinsically embedded in natural languages,  to composite words into sentences for semantic representation \cite{dowty2007compositionality}.
%it seems to have little ability to model the compositionality of language (i.e., syntactic structure).

Some efforts have been made to extend chain-structured LSTMs to tree-structured topologies, i.e., Tree-LSTMs \cite{tai2015improved,zhu2015long}, to address this issue.
Tree-LSTMs process sentences in a hierarchical and recursive manner by propagating information up a given parsing tree to consider  long-distance interactions over syntax structures.
%So the TreeRNNs seem to be the principled model, since meaning in natural language sentences is known to be constructed recursively by a manner of tree structure.
Tree-LSTMs have outperformed chain-structured LSTM-RNNs in some NLP tasks, such as sentiment classification and semantic relatedness analysis \cite{tai2015improved}.
However, there are still some deficiencies with Tree-LSTMs.
First, the construction of Tree-LSTMs depends on external syntax parsers. A high-performance parser is difficult to construct and may not be available for all languages.
Second, Tree-LSTMs derive a state vector for each non-leaf node in the parsing tree. Therefore, combining Tree-LSTMs with other sequential encoders that produce a state vector for each word is not straightforward.
Third, the recursive tree structure makes the training of Tree-LSTMs time consuming
because the batch-mode computation commonly adopted for training conventional LSTM-RNNs cannot be applied to directly train Tree-LSTMs.

%but being largely overlooked in favor of sequence-based RNNs for their incompatibility with batched computation and their reliance on external parsers, which means that training a TreeRNNs model will be very time consuming.
%Anyway, ignoring the time efficiency, the TreeRNNs do not show considerable advantage over RNNs in view of overall practical tasks in NLP, especially for tasks concerning sequence generation, such as machine translation and summarization.

Therefore, this paper proposes a novel neural encoder, named sequential neural encoder with latent structured description (SNELSD), for modeling sentences.
This model introduces latent chunk-level representations into conventional sequential neural encoders, i.e., LSTM-RNNs, to implicitly consider the compositionality of languages in semantic modeling.
Here, word chunks are adopted as intermediate units between words and sentences for sentence modeling.
The boundaries of word chunks are hidden and determined in a task-dependent way, which is different from the conventional text chunking \cite{zhang2002text} task in NLP.
An SNELSD is a hierarchical chain-structured model that is composed of a detection layer and a description layer.
The detection layer estimates the boundaries of latent word chunks and obtains a chunk-level representation for each word.
The description layer processes these representations using modified LSTM units.
The model parameters are estimated in an end-to-end manner without using parsing trees.
Therefore, the outputs of SNELSDs are sequential, and they can be conveniently augmented with other semantic representations, such as word vectors and the states of LSTM-RNNs,
to produce a more comprehensive description of sentence meanings.
%The SNELS model first split a sentence into word blocks and obtain representations of each, then accepts the word blocks as sequential inputs like LSTM.

The proposed SNELSD model has two main characteristics.
First, latent word chunks are adopted as the intermediate units between words and sentences in SNELSD to consider syntax-related structure information during sentence modeling.
The conventional text chunking task aims to identify constituent parts of sentences (nouns, verbs, adjectives, and so forth) and then link them to higher order units (noun groups, verb groups, phrases, and so on),
which is also called shallow parsing.
Chunking sentences into several essential parts is also considered to be a basic cognitive mechanism of human reading.
For example, previous research indicated that chunking reading material or separating sentences into meaningful related parts improved the reading comprehension of readers with low reading ability \cite{casteel1990effects}.
The mental process of chunking words into phrases was considered to be necessary since our mind can not hold  more than approximately four to seven separate items in short-term memory \cite{miller1956magical}.

%The motivation of our designation of SNELS is twofold. On one hand, as we all know, when we read a sentence, especially for a long sentence, we tend to split the sentence into several essential parts, here we call this 'latent chunking', because it's benefit for our grasping of key information and ignorance of insignificant information. Some research results have indicated that 'chunked' reading material or separation of sentences into meaningful related parts will improve reading comprehension of low ability readers\cite{casteel1990effects}, and Miller figured that the mental process of chunking words into phrases is necessary since the mind cannot hold in short-term memory more than approximately four to seven separate items \cite{miller1956magical}. All these show that chunking sentences into essential parts--latent chunking--is a basic cognition process mechanism of human reading. And we can think our computer as a low ability reader, so it's reasonable for our SNELS model to incorporate the 'latent chunking' process.

Second, word chunks are treated as hidden units, and the latent chunk-level representations are embedded in a sequential model structure.
The SNELSD model is designed in an end-to-end manner, which learns to split sentences into hidden word chunks %totally depending on specific tasks
without relying on additional text chunking algorithms.
In other words, the SNELSD model is capable of exploring the latent structure information of sentences in a task-dependent manner.
In contrast to Tree-LSTMs, which adopt tree-structured topologies, the proposed SNELSD model still follows the chain structure,
which guarantees efficient  model training by using batch-mode computation and makes it convenient to combine other sequential sentence encoders.

%On the other hand, in the field of NLP there exists chunking task, which aims to identifies constituent parts of sentences (nouns, verbs, adjectives, etc.) and then links them to higher order units (noun groups, verb groups, phrases, etc.).
%It can serves as a first step for full parsing. Also it's sufficient for many applications, like inforamtion extracion (IE), question answering (QA).
%To a certain extent, chunked sentences is promising to utilize the syntactic structure information. However, chunking, also called shallow parsing, is also a parsing task. So it has the same weakness with those we discussed about TreeRNNs above.
%And we think the 'chunking' pattern while human reading may be more task-dependent than syntax-dependent since human is highly intelligent species.
%Moreover, this can make our SNELS model owns the comparative efficient computation to LSTMs.
% You must have at least 2 lines in the paragraph with the drop letter
% (should never be an issue)

In this paper, we first introduce the architecture and related computational formulas of our SNELSD model after a brief review of related works.
Then, we evaluate the proposed SNELSD model on a natural language inference (NLI) task and a sentiment analysis (SA) task by comparing it with other sentence encoders,
including LSTM-RNNs and Tree-LSTMs.
% in which we compared it with a set of ordinary LSTM networks and TreeLSTM encoding models to analyze the difference and relation among SNELS, LSTMs and TreeLSTM models and compared SNELS models with other published state-of-art results. Also, in order to verify whether our SNELS can capture some regular latent structure information in sentences, we visualized the 'chunking' pattern of SNELS model on NLI task. Further more, in order to verify whether our SNELS model is task-dependent, we conduct the same comparison experiments on sentiment analysis (SA) with that on NLI task and analysed the 'chunking' pattern of SNELS model on SA task.
%In this paper, we first designed the architecture and related computational formulas of SNELS based on ordinary LSTM networks, then we tested it on the Natural Language Inference(NLI)\cite{maccartney2009natural} task based on the infrastructure of Enhanced BiLSTM Inference Model (EBIM)\cite{chen2016enhancing} to decide the best architecture for SNELS model. After that, we compared our best SNELS model with 1-layer or 2-layer LSTM/BiLSTM encoding models to analyze the intrinsic character of SNELS, then we compare SNELS model with other published state-of-art results on NLI task. Further more, in order to verify whether our SNELS model is task-dependent, we conduct the same comparison experiments on sentiment analysis (SA) with that on NLI task and compared the blocking pattern of SNELS  on these two task.
The main contributions of this paper are twofold.
On the one hand, this paper proposes a novel sequential neural encoder that implicitly considers the influence of syntax structure on semantic modeling.
The SNELSD model has a two-layer hierarchical chain structure, which makes a good balance between the flat chain structure (i.e., LSTM-RNN) and the hierarchical tree structure (i.e., Tree-LSTM).
In other words, the SNELSD is an absolutely sequence-based model similar to the LSTM-RNNs,
and it also utilizes the strategy of processing sentences hierarchically as does Tree-LSTMs.
However, our SNELSD model does not rely on additional syntax parsing or text chunking modules
%but task-dependent 'chunking', which we call 'latent chunking' here, so our SNELS exploits a kind of latent structure information.
but rather explores the latent structured information within sentences through end-to-end training.
In our experiments on the NLI and SA tasks, the trained SNELSD models can truly capture some useful and regular chunking patterns that match the intrinsic characteristics of different tasks well.
On the other hand, the proposed SNELSD model helps to obtain better performance than ordinary LSTM-RNNs and Tree-LSTMs on both NLI and SA tasks.
On the Stanford Natural Language Inference (SNLI) task \cite{bowman2015large}, the proposed method achieved an accuracy of $88.3\%$, which is a new state-of-the-art performance without a multi-model ensemble.

%to our knowledge, the SNELS is the first model trying to incorporate the 'chunking' mechanism of human cognition process while reading. In practical test on NLI and SA tasks, the SNELS model can truly capture some useful and regular 'chunking' pattern that matches the intrinsic character of tasks well, which illustrates the flexibility of our SNELS model to be fitted into different tasks. Anyway, the SNELS model can outperform the ordinary LSTMs and Tree-LSTM encoding models on NLI task and do not show weakness than those LSTMs and Tree-LSTM networks on SA task.

\section{Related work}
\subsection{Text Chunking}
Text chunking \cite{zhang2002text}, also called shallow parsing \cite{federici1996shallow}, is an NLP task that aims to identify the constituent parts of sentences (nouns, verbs, adjectives, and so on) and then link them to higher order units (noun groups, verb groups, phrases, and so forth).
%Chunking builds flat structures for sentences not like parsing builds tree structure, it can be much faster, more robust
As an alternative to full parsing, which is more complicated and less robust,
text chunking has been applied to many NLP tasks, such as information extraction (IE) \cite{banko2007open,sarawagi2004semi} and question answering (QA) \cite{soricut2004automatic}, to extract and represent syntax-related information.

The SNELSD model proposed in this paper adopts word chunks as intermediate units between words and sentences for sentence modeling.
The use of word chunks here is different from the conventional text chunking task in two aspects.
First, the conventional text chunking task is generally performed over sequences of part-of-speech tags \cite{ramshaw1995text} based on syntactic regular expressions,
and its outputs typically have syntactic labels.
However, the determination of word chunks in SNELSD models is not syntax dependent but rather task dependent by employing end-to-end model training.
Therefore, the construction of SNELSD models does not rely on additional syntax parsing or text chunking algorithms.
Second, in contrast to text chunking, which provides explicit boundaries of syntactic units, the word chunks in SNELSD models are latent descriptions,
which are represented by the probabilities of  a chunk boundary existing after each word.

%Generally, the 'chunking' procedure is performed over sequences of part-of-speech tags based on syntactic regular expressions. The chief motivation for chunking is to locate information, or to ignore information. So we think in practical tasks, the chunking may not should be syntax-dependent but task-dependent. Hence, the chunking in our SNELS model do not need any syntactic information like part-of-speech tags nor constraint of syntactic regular. We can regard ordinary chunking as a syntax-dependent task, and chunking in SNELS--latent chunking--as a task-dependent but syntax-independent task. Most importantly, this can make our SNELS model owns the comparative efficiency in computation to LSTM.
\begin{figure}[!t]
\centering
\includegraphics[width=2.8in]{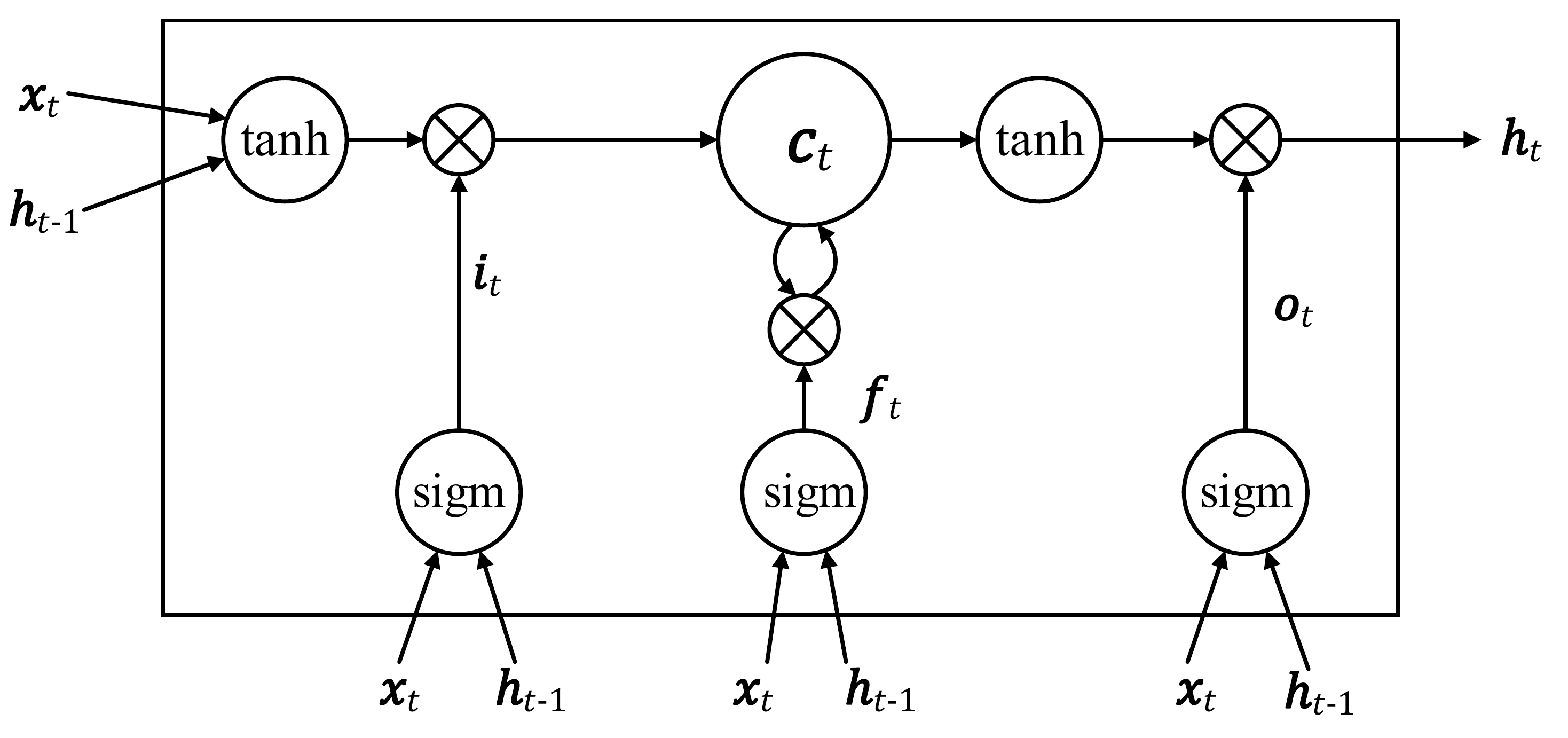}
\caption{Structure of a long short-term memory (LSTM) unit.
}
\label{fig:lstm_unit}
\end{figure}
\subsection{Long Short-Term Memory}
The long short-term memory (LSTM) model \cite{hochreiter1997long} was proposed to address the  gradient vanishing and exploding problem when training recurrent neural networks (RNNs).
RNNs with LSTM cells (LSTM-RNN) have been widely used in NLP applications for the sequential modeling of sentences.
%Hochreiter et Schmidhuber proposed  models  which are widely used in NLP applications. Since our SNELS model is designed based on LSTM model, and our SNELS model is compared with it from factors of hierarchy and direction. So, here we will briefly introduce these ordinary LSTM models.
An LSTM cell is a complex hidden unit.
As shown in Fig. \ref{fig:lstm_unit}, it contains three gates, namely, input gate $\textbf{i}_t$, output gate $\textbf{o}_t$, and forget gate $\textbf{f}_t$, which determine whether to utilize the input, whether to create an output, and whether to update the cell memory state, respectively. Therefore, an RNN that uses LSTM cells is capable of remembering the information from a long span of time steps.
%\subsubsection{LSTM}
%From Fig. \ref{fig:lstm1}, we can know that the core of LSTM networks is the recurrent calculation unit,
The following equations define a regular LSTM unit.
\begin{figure}[!t]
\centering
\subfigure[Flat chain structure of a 1-layer LSTM-RNN.]{
\label{fig:lstm1}
\includegraphics[width=3.4in]{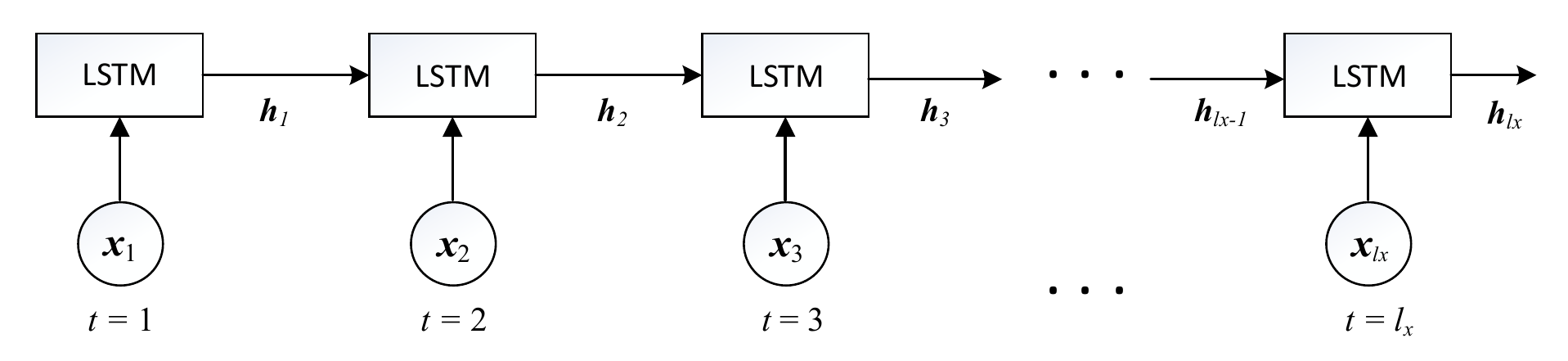}}
\hspace{1in}
\subfigure[Tree structure of a Tree-LSTM.]
{\label{fig:treelstm}
\includegraphics[width=3.0in]{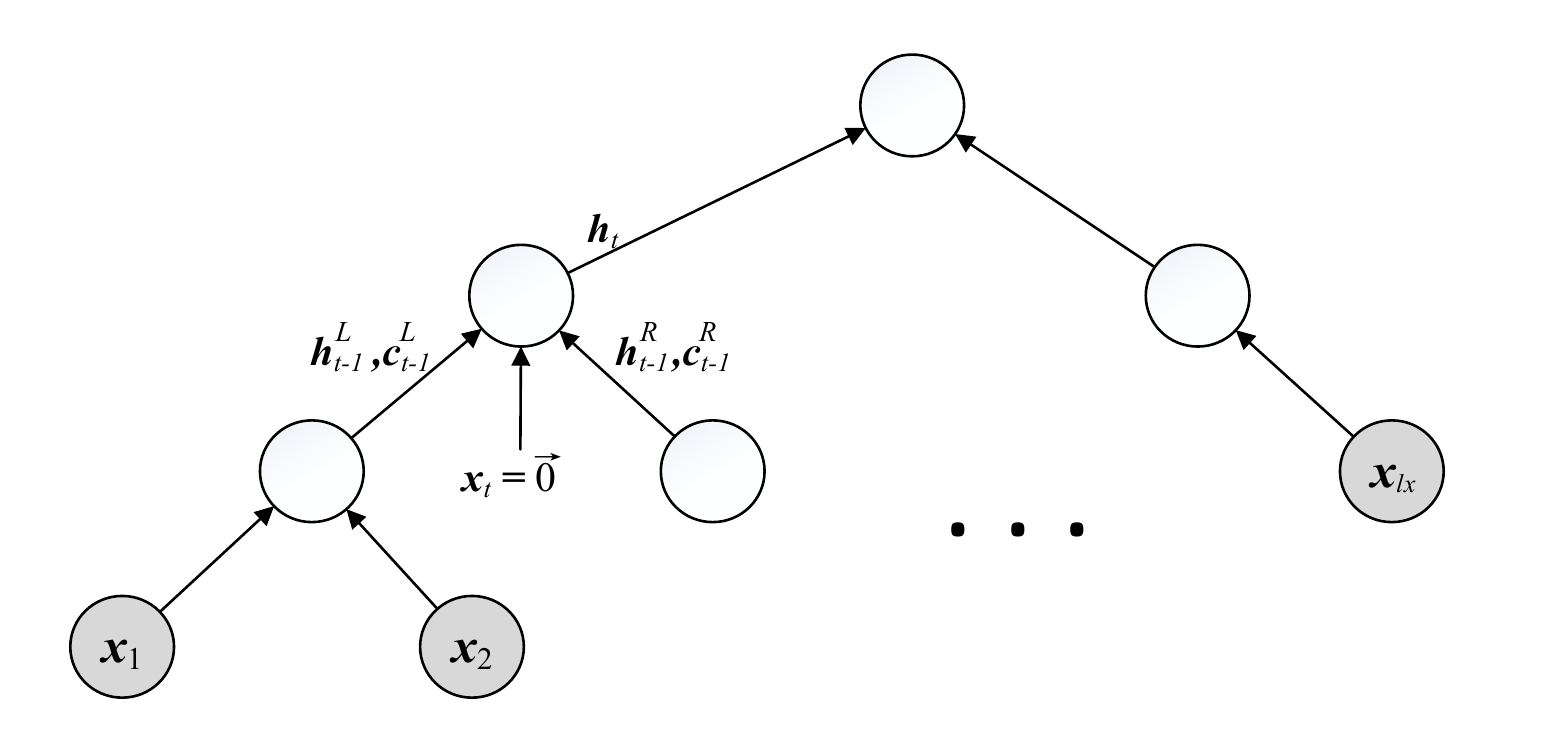}}
\caption{The structures of 1-layer LSTM and Tree-LSTM networks. (a) The flat chain structure of a 1-layer LSTM when processing a sentence with a length of $l_x$.
It accepts a word vector $\textbf{x}_t$ as input and generates a hidden state vector $\textbf{h}_t$ for each word.
(b) The hierarchical recursive structure of a Tree-LSTM.
%It propagates information up a binary parsing tree.
The gray leaf nodes are input word vectors, and the hidden states of all non-leaf nodes can be used to compose the sentence representation.}
\label{fig:lstm}
\end{figure}
%\begin{equation}\label{eq:lstm_1}
%\textbf{i}_t = \sigma (\textbf{W}_i \textbf{x}_t + \textbf{U}_i \textbf{h}_{t-1}),
%\end{equation}
%\begin{equation}\label{eq:lstm_2}
%\textbf{f}_t = \sigma (\textbf{W}_f \textbf{x}_t + \textbf{U}_f \textbf{h}_{t-1}),
%\end{equation}
%\begin{equation}\label{eq:lstm_3}
%\textbf{o}_t = \sigma (\textbf{W}_o \textbf{x}_t + \textbf{U}_o \textbf{h}_{t-1}),
%\end{equation}
%\begin{equation}\label{eq:lstm_4}
%\textbf{c}_t = \textbf{f}_t\odot \textbf{c}_{t-1}+\textbf{i}_t\odot \tanh(\textbf{W}_c \textbf{x}_t + \textbf{U}_c \textbf{h}_{t-1}),
%\end{equation}
%\begin{equation}\label{eq:lstm_5}
%\textbf{h}_t = \textbf{o}_t\odot \textbf{c}_{t},
%\end{equation}
\begin{align}
\textbf{i}_t &= \sigma (\textbf{W}_i \textbf{x}_t + \textbf{U}_i \textbf{h}_{t-1} + \textbf{b}_i),\\
\textbf{f}_t &= \sigma (\textbf{W}_f \textbf{x}_t + \textbf{U}_f \textbf{h}_{t-1} + \textbf{b}_f),\\
\textbf{o}_t &= \sigma (\textbf{W}_o \textbf{x}_t + \textbf{U}_o \textbf{h}_{t-1} + \textbf{b}_o),\\
\textbf{c}_t &= \textbf{f}_t\odot \textbf{c}_{t-1}+\textbf{i}_t\odot \tanh(\textbf{W}_c \textbf{x}_t + \textbf{U}_c \textbf{h}_{t-1}+ \textbf{b}_c),\\
\textbf{h}_t &= \textbf{o}_t\odot \tanh(\textbf{c}_{t}),
\end{align}
where $\sigma$ is the sigmoid function and $\odot$ is the element-wise multiplication between two vectors.
% Note that for brevity, all bias vectors are omitted in this paper.
%(the same simplified handling with all equations description in this Section).
At the $t$-th step, the LSTM unit accepts word vector $\textbf{x}_t$ as input,
and it employs a set of internal vectors, including an input gate $\textbf{i}_t$, a forget gate $\textbf{f}_t$, an output gate $\textbf{o}_t$, and a memory cell $\textbf{c}_t$, to generate a hidden state $\textbf{h}_t$.
Fig. \ref{fig:lstm1} shows the flat chain structure of a 1-layer LSTM-RNN.
For a sentence $\textbf{X}=[\textbf{x}_1, \textbf{x}_2, ..., \textbf{x}_{l_x}]$ with $l_x$ words,
we can use either the sequence  of output hidden states $[\textbf{h}_1, \textbf{h}_2, ..., \textbf{h}_{l_x}]$ or the last hidden state $\textbf{h}_{l_x}$ as the final sentence representation.

To access both the preceding and succeeding contexts, RNNs with bidirectional LSTM units (BLSTM-RNNs) are commonly adopted.
%\subsubsection{BiLSTM}
A BLSTM-RNN can be viewed as two independent LSTM-RNNs that process a sentence along the forward and backward directions.
For example, to encode a sentence $\textbf{X}=[\textbf{x}_1, \textbf{x}_2, ..., \textbf{x}_{l_x}]$ with a length of $l_x$,
the two independent LSTM-RNNs compute the sequences of hidden states $[\overrightarrow{h_1}, \overrightarrow{h_2}, ..., \overrightarrow{h_{l_x}}]$ and $[\overleftarrow{h_1}, \overleftarrow{h_2}, ..., \overleftarrow{h_{l_x}}]$.
Then, these two sequences are merged by concatenating the two hidden states $\overrightarrow{h_t}$ and $\overleftarrow{h_t}$ at the same time position to obtain
%$\left[[\overrightarrow{h_1};\overleftarrow{h_1}], [\overrightarrow{h_2};\overleftarrow{h_2}], ..., [\overrightarrow{h_{l_x}};\overleftarrow{h_{l_x}}]\right]$
the final representation for sentence $\textbf{X}$.

A deep LSTM-RNN can be constructed by stacking multiple recurrent hidden layers one on the top of another.
%\subsubsection{2-layer LSTM}
%It has a two layer hierarchical chain structure, just a LSTM layer piled upon another LSTM layer.
The LSTM units in the first layer accept a word sequence as input, and the upper layers  accept the hidden states of the lower layers as input.
The final sentence representation is composed of the hidden states of the top layer.
A deep BLSTM-RNN can be constructed in a similar way.

Similar to LSTM-RNN, the SNELSD model proposed in this paper is also a sequential encoder for sentence modeling.
However,  the SNELSD adopts a two-layer hierarchical chain structure,
in which the hidden units in the two layers are not LSTMs, and they have different architectures to fulfill different functions.
The details of SNELSDs will be introduced in Section \ref{sec:SNELS}.

%of  $\textbf{S}$ is the hidden state output $(\textbf{h}_1^2, \textbf{h}_2^2, ..., \textbf{h}_{l_x}^2)$ of $LSTM2$.
%\subsubsection{2-layer BiLSTM}
%The model structure of 2-layer BiLSTM is just like 2-layer LSTM, a BiLSTM layer piled upon another BiLSTM layer. Here we do not make more description.
\subsection{Tree-LSTMs}\label{desc:treelstm}
Tree-LSTMs have recently been investigated to incorporate syntactic parsing information for deriving the representation of sentences \cite{tai2015improved,zhu2015long,chen2016enhancing,phong2015compositional}.
Tree-LSTMs are developed from chain-structured LSTMs and have tree-structured network topologies,  as shown in Fig. \ref{fig:treelstm}.
At the $t$-th node of a binary parsing tree, an input vector $\textbf{x}_t$, which is a word vector at leaf nodes and a zero vector at non-leaf nodes, and the hidden vectors of its two child nodes (the left child $\textbf{h}_{t-1}^L$ and the right child $\textbf{h}_{t-1}^L$) are taken as the inputs to calculate the hidden state vector $\textbf{h}_{t}$ of the current node.
These input vectors are used to configure the memory cell $\textbf{c}_t$ and the four gates, i.e., the input gate $\textbf{i}_t$, output gate $\textbf{o}_t$, and two forget gates $\textbf{f}_t^L$ and $\textbf{f}_t^R$.
Moreover, the memory cell $\textbf{c}_t$ considers the memory vectors $\textbf{c}_{t-1}^L$ and $\textbf{c}_{t-1}^R$ from the two child nodes.
Specifically, the forward propagation  of a Tree-LSTM unit can be calculated as follows \cite{chen2016enhancing}.
%\begin{equation}\label{eq:treelstm1}
%\textbf{i}_t = \sigma(\textbf{W}_i \textbf{x}_t + \textbf{U}_i^L \textbf{h}_{t-1}^L + \textbf{U}_i^R \textbf{h}_{t-1}^R),
%\end{equation}
%\begin{equation}\label{eq:treelstm2}
%\textbf{f}_t^L = \sigma(\textbf{W}_f \textbf{x}_t + \textbf{U}_f^{LL} \textbf{h}_{t-1}^L + \textbf{U}_f^{LR} \textbf{h}_{t-1}^R),
%\end{equation}
%\begin{equation}\label{eq:treelstm2}
%\textbf{f}_t^R = \sigma(\textbf{W}_f \textbf{x}_t + \textbf{U}_f^{RL} \textbf{h}_{t-1}^L + \textbf{U}_f^{RL} \textbf{h}_{t-1}^R),
%\end{equation}
%\begin{equation}\label{eq:treelstm3}
%\textbf{o}_t = \sigma(\textbf{W}_o \textbf{x}_t + \textbf{U}_f^L \textbf{h}_{t-1}^L + \textbf{U}_f^R \textbf{h}_{t-1}^R),
%\end{equation}
%\begin{equation}\label{eq:treelstm4}
%\textbf{u}_t = \sigma(\textbf{W}_c \textbf{x}_t + \textbf{U}_c^L \textbf{h}_{t-1}^L + \textbf{U}_c^R \textbf{h}_{t-1}^R),
%\end{equation}
%\begin{equation}\label{eq:treelstm5}
%\textbf{c}_t = \textbf{f}_t^L \odot \textbf{c}_{t-1}^L + \textbf{f}_t^R \odot \textbf{c}_{t-1}^R + \textbf{i}_t \odot \textbf{u}_t,
%\end{equation}
%\begin{equation}\label{eq:treelstm6}
%\textbf{h}_t = \textbf{o}_t \odot \tanh(\textbf{c}_t).
%\end{equation}
\begin{align}
\textbf{i}_t &= \sigma(\textbf{W}_i \textbf{x}_t + \textbf{U}_i^L \textbf{h}_{t-1}^L + \textbf{U}_i^R \textbf{h}_{t-1}^R + \textbf{b}_i),\\
\textbf{f}_t^L &= \sigma(\textbf{W}_f \textbf{x}_t + \textbf{U}_f^{LL} \textbf{h}_{t-1}^L + \textbf{U}_f^{LR} \textbf{h}_{t-1}^R + \textbf{b}_f^L),\\
\textbf{f}_t^R &= \sigma(\textbf{W}_f \textbf{x}_t + \textbf{U}_f^{RL} \textbf{h}_{t-1}^L + \textbf{U}_f^{RR} \textbf{h}_{t-1}^R + \textbf{b}_f^R),\\
\textbf{o}_t &= \sigma(\textbf{W}_o \textbf{x}_t + \textbf{U}_o^L \textbf{h}_{t-1}^L + \textbf{U}_o^R \textbf{h}_{t-1}^R + \textbf{b}_o),\\
\textbf{u}_t &= \sigma(\textbf{W}_c \textbf{x}_t + \textbf{U}_c^L \textbf{h}_{t-1}^L + \textbf{U}_c^R \textbf{h}_{t-1}^R + \textbf{b}_u),\\
\textbf{c}_t &= \textbf{f}_t^L \odot \textbf{c}_{t-1}^L + \textbf{f}_t^R \odot \textbf{c}_{t-1}^R + \textbf{i}_t \odot \textbf{u}_t,\\
\textbf{h}_t &= \textbf{o}_t \odot \tanh(\textbf{c}_t).
\end{align}
%In our experiments, we used binary tree-LSTM, the tree structure for each sentence is produced by a constituency parse (here we used the stanford-parser\cite{}).
%In our experiments, we compared the SNELS model with Tree-LSTM since SNELS utilize the latent structure information not like Tree-LSTM utilizling the syntactic structure information.

Similar to Tree-LSTMs, the SNELSD model proposed in this paper also aims to introduce the compositionality of languages into the semantic modeling of sentences.
However, the SNELSD model utilizes hidden word chunks and an end-to-end training strategy, thus avoiding the reliance on additional syntax parsing.
Furthermore, SNELSD is a sequential model, not a recursive one with tree structures, which guarantees the efficiency of model training and the convenience of combining other sequential encoders.

\begin{figure}[!t]
\centering
\includegraphics[width=3.6in]{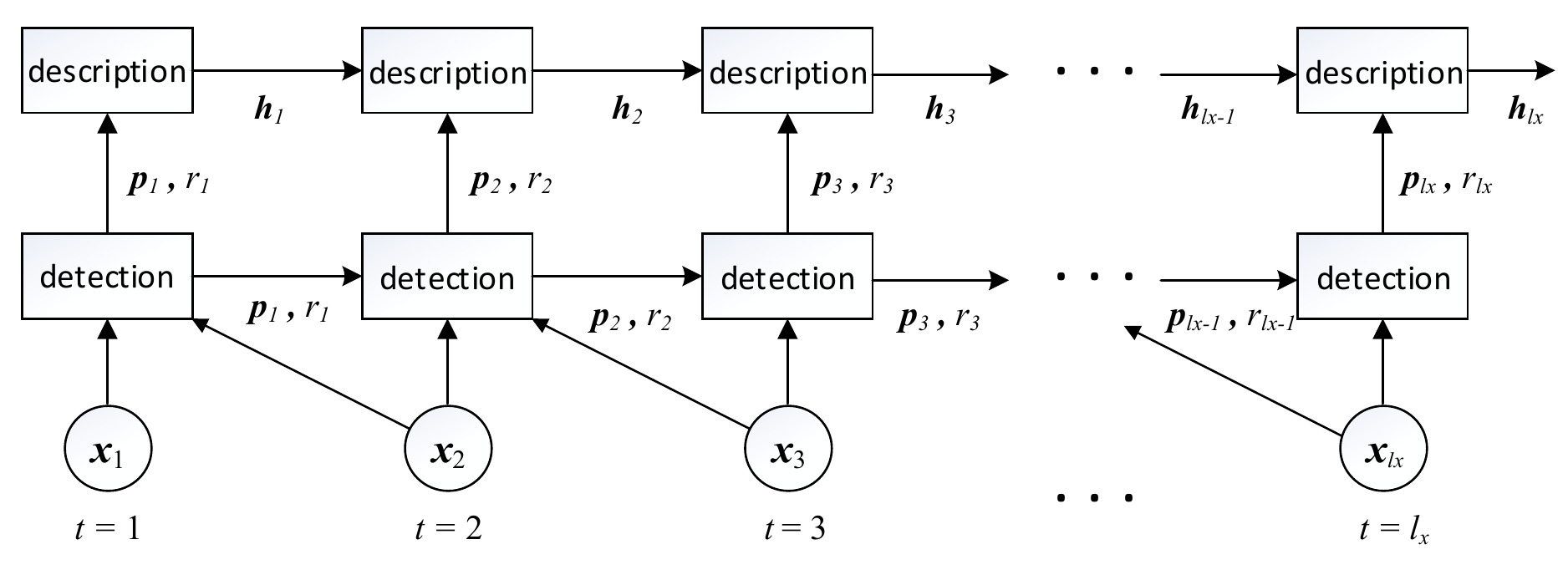}
\caption{The two-layer hierarchical chain structure of an SNELSD.
The first layer, the detection layer, accepts the  word vector sequence  $[\textbf{x}_1, \textbf{x}_2, ..., \textbf{x}_{l_x}]$ as input
and produces a sequence of chunk-level vector representations $[\textbf{p}_1, \textbf{p}_2, ..., \textbf{p}_{l_x}]$ together with a sequence of boundary indicators $[r_1, r_2, ..., r_{l_x}]$.
The second  layer, the description layer, accepts the outputs of the detector layer and produces sequential encoding outputs.
}
\label{fig:SNELS}
\end{figure}
\section{Sequential Neural Encoder with Latent Structured Description (SNELSD)}\label{sec:SNELS}
The LSTM-RNN and Tree-LSTM models exploit a flat chain structure and a hierarchical tree structure, respectively.
Our SNELSD model utilizes a hierarchical chain structure, which can be considered as a trade-off between LSTM-RNNs and Tree-LSTMs.
Specifically, an SNELSD model has a two-layer structure as shown in Fig. \ref{fig:SNELS}.
%The first layer, named \textbf{detector layer} here, accepts the word vector sequence as input and splits the sentence into word blocks, then derive the vector representation for each word block. The second layer, we call \textbf{descriptor layer} here, accepts the sequence of word block representation as input and derive the final contextual representation for sentences.
The first layer, named the \emph{detection layer}, predicts the  boundaries of latent word chunks in an input sentence and derives a chunk-level vector for each word.
The second layer, named the \emph{description layer}, utilizes modified LSTM units to process these chunk-level vectors in a recurrent manner and produces sequential encoding outputs.
The details of these two layers will be introduced in this section.
%In order to make our SNELS own the comparative compatibility with batch computation compared to LSTM, we should ensure both detector layer and descriptor layer work in a rigorous sequential mechanism like the one in LSTM. Specifically, for example, the detector layer should output a word block sequence with the same length of word sequence it accepted.
\subsection{Detection Layer}
\begin{figure}[b]
\centering
\includegraphics[width=2.8in]{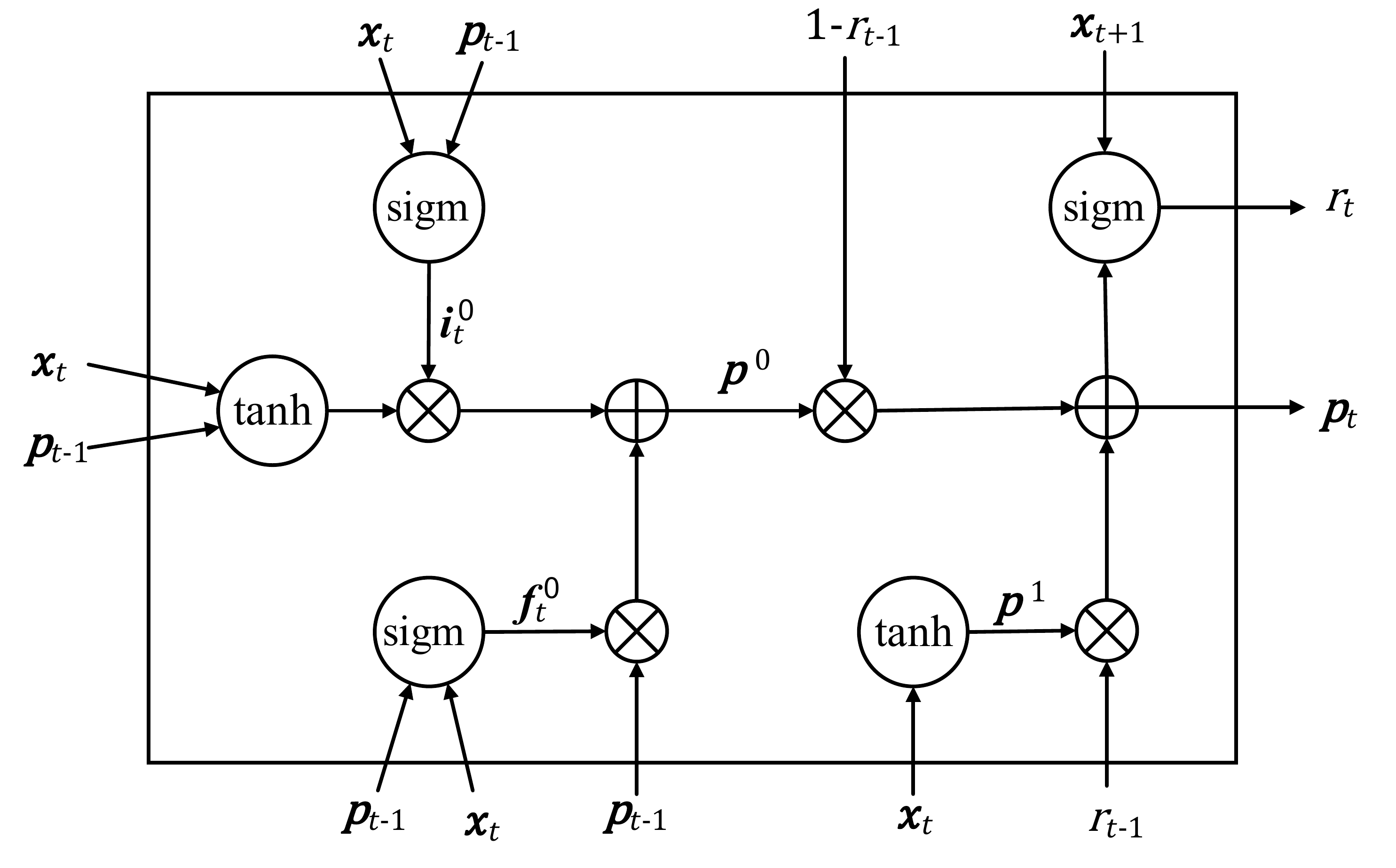}
\caption{Structure of a detection unit in the proposed SNELSD model.
%At each time step $t$, it accepts the word vectors $\textbf{x}_t$ and $\textbf{x}_{t+1}$ and boundary indicator coefficient $r_{t-1}$ as inputs, then outputs the vector representation of word block $\textbf{p}_t$ and the corresponding boundary indicator coefficient $r_t$
}
\label{fig:dector}
\end{figure}
The detection layer is designed to segment a sentence into word chunks.
For a sentence %$\textbf{S}:(\textbf{x}_1, \textbf{x}_2, ..., \textbf{x}_{l_x})$
with $l_x$ words, the detection layer accepts the sequence of word vectors $[\textbf{x}_1, \textbf{x}_2, ..., \textbf{x}_{l_x}]$ as input
and produces a sequence of chunk-level vector representations $[\textbf{p}_1, \textbf{p}_2, ..., \textbf{p}_{l_x}]$ together with a sequence of boundary indicators $[r_1, r_2, ..., r_{l_x}]$.
The value $r_t$ represents the possibility of  a chunk boundary existing after word $\textbf{x}_t$, which is a continuous measurement between $0.0$ and $1.0$.
%The higher is the $r_t$, the greater the possibility will be.
In other words, the word chunks are considered to be hidden units in the proposed SNELSD model.

The structure of a detection unit is shown in Fig. \ref{fig:dector}.
At the $t$-th step, the detection unit receives the current word vector $\textbf{x}_t$, the next word vector $\textbf{x}_{t+1}$, the chunk boundary indicator of the previous step $r_{t-1}$ and the chunk-level representation of the previous step $\textbf{p}_{t-1}$ as input,
and it returns a chunk boundary indicator $r_t$ together with the updated chunk-level representation at the current step $\textbf{p}_t$.
Mathematically, a detection unit is defined by the following equations. %Note that for brevity, all bias vectors are omitted (the same simplified handling with all equations description in this Section).
%\begin{equation}\label{eq:soft_block1}
%\textbf{i}_t^0 = \sigma(\textbf{W}_i^0 \textbf{x}_t + \textbf{U}_i^0 \textbf{p}_{t-1}),
%\end{equation}
%\begin{equation}\label{eq:soft_block2}
%\textbf{f}_t^0 = \sigma(\textbf{W}_f^0 \textbf{x}_t + \textbf{U}_f^0 \textbf{p}_{t-1}),
%\end{equation}
%\begin{equation}\label{eq:soft_block3}
%\textbf{p}^0 = \textbf{f}_t^0\odot \textbf{p}_{t-1} + \textbf{i}_t^0\odot \tanh(\textbf{W}_p^0 \textbf{x}_t + \textbf{U}_p^0 \textbf{p}_{t-1}),
%\end{equation}
%\begin{equation}\label{eq:soft_block4}
%\textbf{p}^1 = \tanh(\textbf{W}_p^1 \textbf{x}_t),
%\end{equation}
%\begin{equation}\label{eq:soft_block5}
%\textbf{p}_t = (1-r_{t-1})*\textbf{p}^0 + r_{t-1}*\textbf{p}^1,
%\end{equation}
%\begin{equation}\label{eq:soft_block6}
%{r}_t = \sigma({[\textbf{p}_t;\textbf{x}_{t+1}]}^T \textbf{u}_r),
%\end{equation}
\begin{align}
\textbf{i}_t^0 &= \sigma(\textbf{W}_i^0 \textbf{x}_t + \textbf{U}_i^0 \textbf{p}_{t-1} + \textbf{b}_i^0),\\
\textbf{f}_t^0 &= \sigma(\textbf{W}_f^0 \textbf{x}_t + \textbf{U}_f^0 \textbf{p}_{t-1} + \textbf{b}_f^0),\\
\textbf{p}^0 &= \textbf{f}_t^0\odot \textbf{p}_{t-1} + \textbf{i}_t^0\odot \tanh(\textbf{W}_p^0 \textbf{x}_t + \textbf{U}_p^0 \textbf{p}_{t-1} + \textbf{b}_p^0),\\
\textbf{p}^1 &= \tanh(\textbf{W}_p^1 \textbf{x}_t + \textbf{b}_p^1),\\
\textbf{p}_t &= (1-r_{t-1})*\textbf{p}^0 + r_{t-1}*\textbf{p}^1,\\
{r}_t &= \sigma({[\textbf{p}_t;\textbf{x}_{t+1}]}^T \textbf{u}_r).
\end{align}
%The acquisition of $\textbf{p}_t$ can be divided into 3 steps:

These equations can be divided into three operations.
The first is when the words $\textbf{x}_{t-1}$ and $\textbf{x}_{t}$ belong to the same word chunk, which means that %$r_{t-1}$ is $0$, in other words,
the word $\textbf{x}_{t}$ is not at the beginning of a new word chunk.
The updated chunk-level representation, denoted as $\textbf{p}^0$ here, depends on both the input word $\textbf{x}_t$ and the previous chunk-level representation $\textbf{p}_{t-1}$, as shown in (13)-(15).
Two internal gate vectors, i.e., an input gate $\textbf{i}_t^0$ and a forget gate $\textbf{f}_t^0$, are adopted to control the continuous encoding process within a certain word chunk.
This is similar to the conventional LSTM unit since a word chunk can be considered as a very short sentence.
Second, when the word $\textbf{x}_{t-1}$ is at the end of the last word chunk and the word $\textbf{x}_{t}$ is at the beginning of a new word chunk,
the updated chunk-level representation, denoted as $\textbf{p}^1$ here, only depends on the input word $\textbf{x}_t$, as shown in (16).
Finally,  these two different chunk-level representations $\textbf{p}^0$ and $\textbf{p}^1$ are linearly fused using the boundary indicator $r_{t-1}$, as shown in (17).
The boundary indicator $r_{t}$, which describes the possibility of  a chunk boundary existing after word $\textbf{x}_t$, is also calculated using $\textbf{p}_t$ and the following word $\textbf{x}_{t+1}$, as shown in (18).

\subsection{Description Layer}
\begin{figure}[b]
\centering
\includegraphics[width=2.8in]{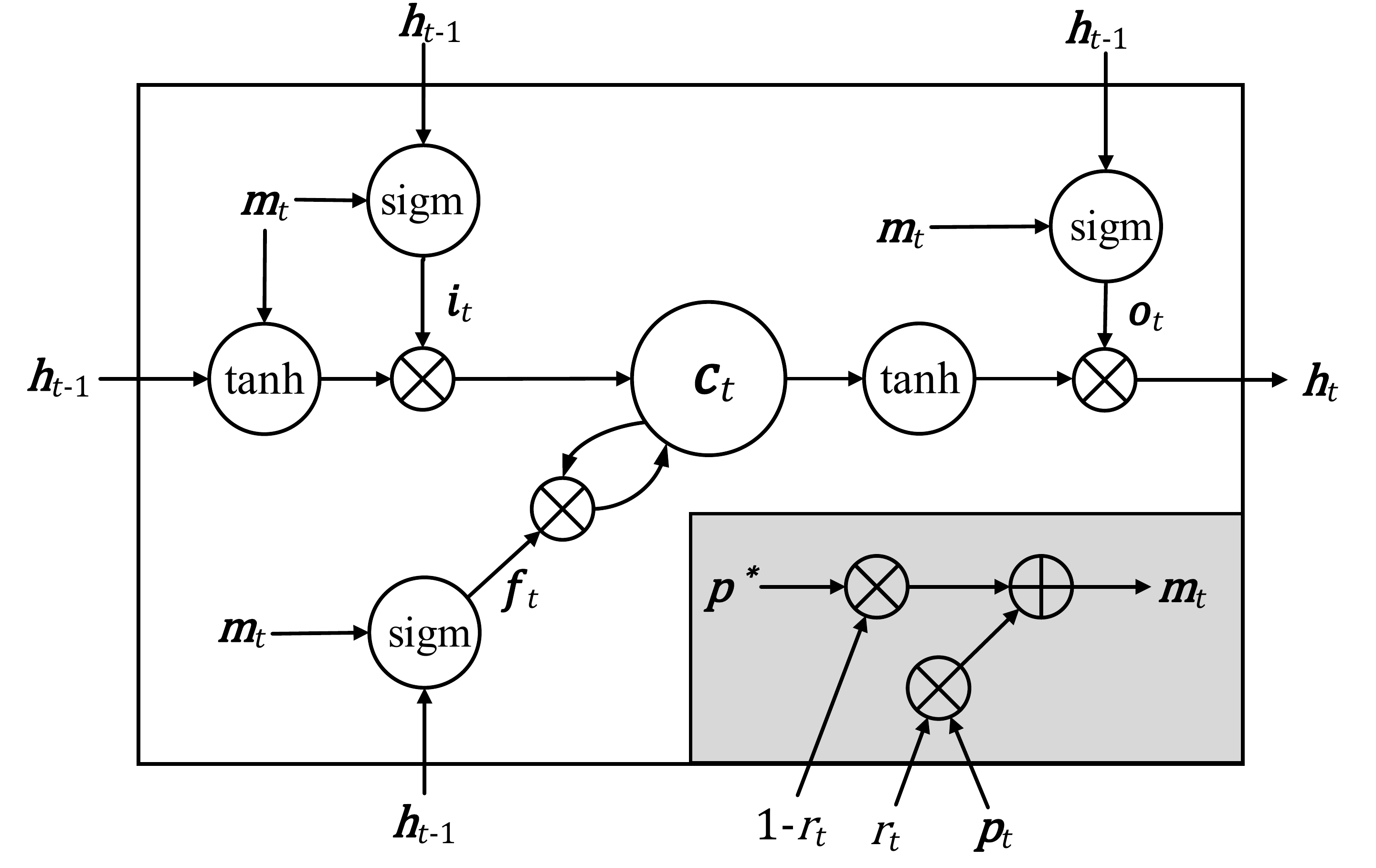}
\caption{Structure of a description unit in the proposed SNELSD model.
%At each time step $t$, it accepts word block representation $\textbf{p}_t$ and boundary indicator coefficient $r_t$ as inputs, then outputs the hidden state $\textbf{h}_t$, which can be used as the contextual representation of sentence $\textbf{S}$
}
\label{fig:descriptor}
\end{figure}
The function of the description layer is to derive the representation of a sentence from the sequence of chunk-level representations $[\textbf{p}_1, \textbf{p}_2, ..., \textbf{p}_{l_x}]$ given by the detection layer.
The structure of a description unit is shown in Fig. \ref{fig:descriptor} and is formulated as follows.

%From the description of detector layer above, we can know that all items in word block representations $(\textbf{p}_1, \textbf{p}_2, ..., \textbf{p}_{l_x})$ can be used to represent a word block in a probabilistic fashion. So the descriptor unit can not handle all items in $(\textbf{p}_1, \textbf{p}_2, ..., \textbf{p}_{L_x})$ samely.
%The descriptor unit is defined by following equations,
%\begin{equation}\label{eq:fitting1}
%\textbf{m}_t = (1-r_t)*\textbf{inp}_{non} + r_t * \textbf{p}_t,
%\end{equation}
%\begin{equation}\label{eq:fitting2}
%\textbf{i}_t = \sigma(\textbf{U}_i \textbf{h}_{t-1} + \textbf{W}_i \textbf{m}_t),
%\end{equation}
%\begin{equation}\label{eq:fitting3}
%\textbf{f}_t = \sigma(\textbf{U}_f \textbf{h}_{t-1} + \textbf{W}_f\textbf{m}_t),
%\end{equation}
%\begin{equation}\label{eq:fitting4}
%\textbf{o}_t = \sigma(\textbf{U}_o \textbf{h}_{t-1} + \textbf{W}_o\textbf{m}_t),
%\end{equation}
%\begin{equation}\label{eq:fitting5}
%\textbf{c}_t = \textbf{f}_t \odot \textbf{c}_{t-1} + \textbf{i}_t \odot\tanh(\textbf{U}_c \textbf{h}_{t-1} + \textbf{W}_c\textbf{m}_t),
%\end{equation}
%\begin{equation}\label{eq:fitting6}
%\textbf{h}_t = \textbf{o}_t \odot \textbf{c}_{t},
%\end{equation}
\begin{align}
\textbf{m}_t &= (1-r_t)*\textbf{p}^* + r_t * \textbf{p}_t,\\
\textbf{i}_t &= \sigma(\textbf{U}_i \textbf{h}_{t-1} + \textbf{W}_i \textbf{m}_t + \textbf{b}_i),\\
\textbf{f}_t &= \sigma(\textbf{U}_f \textbf{h}_{t-1} + \textbf{W}_f\textbf{m}_t + \textbf{b}_f),\\
\textbf{o}_t &= \sigma(\textbf{U}_o \textbf{h}_{t-1} + \textbf{W}_o\textbf{m}_t + \textbf{b}_o),
\end{align}
\begin{align}
\textbf{c}_t &= \textbf{f}_t \odot \textbf{c}_{t-1} + \textbf{i}_t \odot\tanh(\textbf{U}_c \textbf{h}_{t-1} + \textbf{W}_c\textbf{m}_t + \textbf{b}_c),\\
\textbf{h}_t &= \textbf{o}_t \odot \tanh(\textbf{c}_{t}).
\end{align}
At the $t$-th step, the description unit accepts the chunk-level representation $\textbf{p}_t$, the boundary indicator $\textbf{r}_t$ and the hidden state of the previous step  $\textbf{h}_{t-1}$ as inputs and generates an updated hidden state $\textbf{h}_t$.
A description unit is a modified LSTM unit.
The difference is that a blended input vector $\textbf{m}_t$ is calculated to replace $\textbf{p}_t$ for sequential encoding, as shown in (19).
In this equation, $\textbf{p}^*$ is a constant vector to be estimated.
The motivation for introducing $\textbf{m}_t$  is to emphasize the chunk-level representation $\textbf{p}_t$ of the words, which is the last word of a chunk (i.e., $r_t=1$),
and to ignore the chunk-level representation of intermediary words within a chunk (i.e., $r_t=0$).
%and linearly mixes the $\textbf{p}_t$ with $\textbf{inp}_{non}$ using $\textbf{r}_t$ as the weight factor. Then the mixed result $\textbf{m}_t$, we call \emph{blend input} here, can be viewed as the extrinsic motivation, acts the role that changing hidden state $\textbf{h}_t$. Here, the variable $\textbf{inp}_{non}$ is used as a substitute input when $\textbf{r}_t = 0$, the reason for this is that we hope the descriptor layer can output nearly the same hidden state $\textbf{h}_t$ with $\textbf{h}_{t-1}$ when word $\textbf{x}_t$ is not the end of word block.

From (16)-(24), we find that the detector layer  degrades to a simple $tanh$ project layer to process word vectors and the description unit is simply identical to a conventional LSTM unit  when $r_t$ is constantly equal to $1$.

An SNELSD model can work in either \emph{stand-alone mode} or \emph{joint mode}.
In stand-alone mode, the sequence of hidden state vectors $[\textbf{h}_1, \textbf{h}_2, ..., \textbf{h}_{l_x}]$ produced by the description layer are used directly for sentence representation.
In joint mode, these hidden state vectors are further concatenated with word vectors $[\textbf{x}_1, \textbf{x}_2, ..., \textbf{x}_{l_x}]$ or the output of an auxiliary chain-structured sentence encoder (e.g., BLSTM-RNN) at each step to produce a more comprehensive description of sentence meanings.
In either mode, all the model parameters including word embeddings and all transformation matrices are learned in an end-to-end  manner.

\section{Evaluation Tasks}
Two NLP tasks, namely, natural language inference (NLI) and sentiment analysis (SA), are adopted to evaluate the performance of our proposed SNELSD model.
In this section, we briefly introduce these two tasks and explain how to apply SNELSD models to accomplish these two tasks.

\subsection{Natural Language Inference (NLI)} \label{nli_descripttion}
Understanding entailment and contradiction is fundamental to understanding the meaning of natural language \cite{maccartney2008modeling, bowman2015large}.
Thus, the inference about entailment and contradiction is a valuable task  for evaluating sentence encoding models.
Specifically, the natural language inference (NLI) task aims to  determine whether a natural language hypothesis $h$ can be inferred from a natural language premise $p$, as described in the following example \cite{maccartney2008modeling}.\\
\\
$p$: \emph{Several airlines polled saw costs grow more than expected, even after adjusting for inflation.}\\
$h$: \emph{Some of the companies in the poll reported cost increases.}\\ \\
In this example, the hypothesis $h$ can be regarded as being entailed from the premise $p$.

In recent years, there have been advances in NLI.
One main advancement is the availability of a large annotated dataset, the Stanford Natural Language Inference (SNLI) dataset \cite{bowman2015large}, which contains 570K human-written English sentence pairs manually labeled by multiple human subjects.
For each sentence pair, the annotation indicates whether the premise sentence entails the hypothesis sentence, whether they are contradicting each other or whether they have no inference relation.
With its large-scale and human-grounded annotation, the SNLI dataset is competent for training rather complicated sentence encoding models.
%In this benchmark dataset, the sentence encoding models need to decide if a premise $p$ entails a hypothesis $h$, if they are contradicting to each other or have no inference relation.
%\subsection{architecture of testing on NLI}

We evaluate our SNELSD model based on the infrastructure of the enhanced BLSTM inference model (EBIM) \cite{chen2016enhancing}, which has achieved nearly state-of-the-art performance on the SNLI task.
Let two sentences of word vectors $\textbf{A} = [\textbf{a}_1, \textbf{a}_2, ..., \textbf{a}_{l_a}]$ and $\textbf{B} = [\textbf{b}_1, \textbf{b}_2, ..., \textbf{b}_{l_b}]$ denote the premise sentence and the hypothesis sentence.
Each $\textbf{a}_i$ or $\textbf{b}_j \in \mathbb{R}^l$ is an embedding of an $l$-dimensional vector, which can be initialized with pre-trained word embeddings.
The goal is to predict a label $y$ that indicates the inference relationship between $\textbf{A}$ and $\textbf{B}$.
The architecture of EBIM can be divided into four parts, as follows.
\subsubsection{Sentence encoding}
First, the premise and hypothesis sentences $\textbf{A}$ and $\textbf{B}$ are input into a sentence encoder to obtain their context representation vectors $[\bar{\textbf{a}}_1, ..., \bar{\textbf{a}}_{n_a}]$ and $[\bar{\textbf{b}}_1, ..., \bar{\textbf{b}}_{n_b}]$ as
\begin{align}\label{eq:nli1}
\bar{\textbf{a}}_i &= \textbf{Sentence Encoder}(\textbf{A}), \forall i \in [1, ..., n_a],\\
%\end{equation}
%\begin{equation}\label{eq:nli2}
\bar{\textbf{b}}_j &= \textbf{Sentence Encoder}(\textbf{B}), \forall j \in [1, ..., n_b].
\end{align}
In the original EBIM, the sentence encoder is a single-layer BLSTM-RNN model.
Here, we replace it with the other models introduced above for comparison, including multi-layer LSTM-RNN and BLSTM-RNN, Tree-LSTM, and the SNELSD model proposed in this paper.
Note that when a Tree-LSTM is utilized to encode sentences,
the number of derived representation vectors $n_a$ and $n_b$ is equal to the number of non-leaf nodes in the binary syntactic trees of these two sentences.
Otherwise, $n_a=l_a$ and $n_b=l_b$ because all the other models have sequential structures.
%, specifically, all the $l_a$ and $l_b$ in equations (\ref{eq:nli1}-\ref{eq:nli16}) will be replaced with $n_a$ and $n_b$, where $n_a$ and $n_b$ means the number of non-leaf nodes in binary syntactic tree of sentence $\textbf{A}$ and $\textbf{B}$ respectively.
\subsubsection{Soft alignment}
The attention weights $e_{ij}$ between each pair of $\bar{\textbf{a}}_i$ and $\bar{\textbf{b}}_j$ are calculated as
\begin{equation}\label{eq:nli3}
e_{ij} = \bar{\textbf{a}}_i^\top \bar{\textbf{b}}_j, \forall i \in [1, ..., n_a], \forall j \in [1, ..., n_b].
\end{equation}
Subsequently, the attention weights $e_{ij}$ are normalized and are utilized to obtain soft alignment vectors $\tilde{\textbf{a}}_i$ and $\tilde{\textbf{b}}_j$ for both sentences as
\begin{align}\label{eq:nli4}
\tilde{\textbf{a}}_i &= \sum_{j=1}^{l_b} \frac{\exp{e_{ij}}}{\sum_{k=1}^{l_b} \exp{e_{ik}}} \bar{\textbf{b}}_j, \forall i \in [1, ..., n_a],\\
%\end{equation}
%\begin{equation}\label{eq:nli5}
\tilde{\textbf{b}}_j &= \sum_{i=1}^{l_a} \frac{\exp{e_{ij}}}{\sum_{k=1}^{l_a} \exp{e_{ik}}} \bar{\textbf{a}}_i, \forall j \in [1, ..., n_b].
\end{align}
\subsubsection{Inference collection and composition}
We collect the inference-related information using the context representation vectors $\bar{\textbf{a}}_i, \bar{\textbf{b}}_j$ and soft alignment vectors $\tilde{\textbf{a}}_i, \tilde{\textbf{b}}_j$ calculated above.
Specifically, the EBIM model uses vector concatenation, vector difference, and element-wise product to collect the inference sequences $\textbf{m}_a$ and $\textbf{m}_b$ as follows:
\begin{equation}\label{eq:nli6}
\textbf{m}_a = [\bar{\textbf{a}}; \tilde{\textbf{a}}; \bar{\textbf{a}}-\tilde{\textbf{a}}; \bar{\textbf{a}}\odot\tilde{\textbf{a}}],
\end{equation}
\begin{equation}\label{eq:nli7}
\textbf{m}_b = [\bar{\textbf{b}}; \tilde{\textbf{b}}; \bar{\textbf{b}}-\tilde{\textbf{b}}; \bar{\textbf{b}}\odot\tilde{\textbf{b}}],
\end{equation}
where $\bar{\textbf{a}}, \bar{\textbf{b}},\tilde{\textbf{a}}$, and $\tilde{\textbf{b}}$ are the sequences composed of $\bar{\textbf{a}}_i, \bar{\textbf{b}}_j, \tilde{\textbf{a}}_i$, and $\tilde{\textbf{b}}_j$, respectively.
%\subsubsection{Inference Composition}
Then, the sequences $\textbf{m}_a$ and $\textbf{m}_b$ are composed using a BLSTM-RNN model together with average and max pooling to obtain a fixed-dimensional discriminative vector $\textbf{v}$, which describes the inference relationship between the entire premise sentence and its hypothesis.
The calculation is as follows:
\begin{equation}\label{eq:nli8}
\textbf{v}_{1,i} =  \textbf{BLSTM}(\textbf{m}_a), \forall i \in [1, ..., n_a],
\end{equation}
\begin{equation}\label{eq:nli9}
\textbf{v}_{2,j} =  \textbf{BLSTM}(\textbf{m}_b), \forall j \in [1, ..., n_b],
\end{equation}
%Then we convert the resulting vectors obtained above to a fixed-length vector. Here we use average pooling and max pooling technology and concatenate these vectors to get fixed length vector $\textbf{v}$.
\begin{equation}\label{eq:nli10}
\textbf{v}_{1,ave} = \sum_{i=1}^{n_a} \textbf{v}_{1,i}/n_a,\quad\textbf{v}_{1,max} = \max_{i=1}^{n_a} \textbf{v}_{1,i},
\end{equation}
\begin{equation}\label{eq:nli11}
\textbf{v}_{2,ave} = \sum_{j=1}^{n_b} \textbf{v}_{2,j}/n_b,\quad\textbf{v}_{2,max} = \max_{j=1}^{n_b} \textbf{v}_{2,j},
\end{equation}
\begin{equation}\label{eq:nli14}
\textbf{v} = [\textbf{v}_{1,ave}; \textbf{v}_{2,ave}; \textbf{v}_{1,max}; \textbf{v}_{2,max}].
\end{equation}
\subsubsection{Inference determination}
Finally, the vector $\textbf{v}$ is fed into a multi-layer perceptron (MLP) classifier that has a hidden layer with \emph{tanh} activation and a \emph{softmax} output layer.
%\begin{equation}\label{eq:nli15}
%\textbf{h} = \tanh(\textbf{W}_1\textbf{v}+b_1),
%\end{equation}
%\begin{equation}\label{eq:nli16}
%\textbf{y} = softmax(\textbf{W}_2\textbf{h}+b_2),
%\end{equation}
The output vector $\textbf{y}$ is a 3-dimensional vector that indicates the probability of three different inference relationships between the premise sentence and its hypothesis.

\subsection{Sentiment Analysis (SA)}
Sentiment analysis (SA) \cite{pang2008opinion} aims to determine the attitude of someone with respect to some topics or the overall contextual polarity or emotional reaction.
There are many commercial and social applications related to sentiment analysis.
Actually, sentiment analysis can utilize multimodal data including text, speech and video %, and multimodal sentiment analysis is a developing area of research
\cite{DBLP:conf/emnlp/ZadehCPCM17,DBLP:conf/icdm/PoriaCCH16,DBLP:conf/acl/PoriaCHMZM17}.
%Hence, sentiment analysis can also be a good task to evaluate the sentence encoding models.
In natural language processing, a basic sentiment analysis task is to classify the polarity of a given text at the document, sentence, or feature and aspect level.
Therefore, sentence modeling is essential for this task.
In this paper, we evaluate our SNELSD model on the sentiment polarity task using the Stanford Sentiment Treebank dataset (SST) \cite{socher2013recursive}.
This dataset includes fine-grained sentiment labels for 215,154 phrases in the parse trees of 11,855 sentences.
Each label can be one of the 5 sentiment classes from very negative to very positive ($--,-,0,+,++$). The following examples are taken from this dataset, in which only the sentence-level labels are shown.\\
\\
$+$: \emph{taking care of my cat offers a refreshingly different slice of Asian cinema }\\
$-$: \emph{no movement, no yuks, not much of anything}\\\\
%In our experiments, we do not use the phrase-level sentiment label on account of our SNELS model will learn to split the sentence into word blocks, so we exclude the phrases from the training process of SNELS.
Since the focus of this paper is sentence modeling, we exclude the phrase-level samples in the SST dataset and only use sentence-level samples for evaluating different sentence encoders.

%\subsection{architecture of testing on SA}
The overall architecture of the SA model used in this paper is much simpler than that for NLI.
In an input sentence $\textbf{X} = [\textbf{x}_1, \textbf{x}_2, ..., \textbf{x}_{l_x}]$, each $\textbf{x}_i \in \mathbb{R}^l$ is an $l$-dimensional embedding vector, which can be initialized with pre-trained word embeddings.
%The training goal of the model is to predict the sentiment label for sentence $\textbf{X}$. The detailed steps are defined as follows:
First, the sequence of words $\textbf{X}$ is sent into a sentence encoder to obtain the context representations $[\bar{\textbf{x}}_1, ..., \bar{\textbf{x}}_{n_x}]$ as
\begin{equation}\label{eq:sa1}
\bar{\textbf{x}}_i = \textbf{Sentence Encoder}(\textbf{X}), \forall i \in [1, ..., n_x].
\end{equation}
Here, the sentence encoder can be LSTM-RNN or the proposed SNELSD model.
Similar to (25) and (26),  $n_x=l_x$ since both LSTM-RNN and SNELSD model has sequential structures.
Then, a discriminative vector $\textbf{v}$ is obtained by performing average and max pooling on $[\bar{\textbf{x}}_1, ..., \bar{\textbf{x}}_{n_x}]$ as
\begin{align}\label{eq:sa2}
\textbf{v}_{ave} &= \sum_{i=1}^{n_x} \bar{\textbf{x}}_i/n_x, \quad \textbf{v}_{max} = \max_{i=1}^{n_x} \bar{\textbf{x}}_i,\\
%\end{equation}
%\begin{equation}\label{eq:sa4}
\textbf{v} &= [\textbf{v}_{ave}; \textbf{v}_{max}].
\end{align}
%\begin{equation}\label{eq:sa5}
%\textbf{h} = Relu(\textbf{W}_1 \textbf{v}+b_1),
%\end{equation}
%\begin{equation}\label{eq:sa6}
%\textbf{y} = softmax(\textbf{W}_2\textbf{h}+b_2),
%\end{equation}
Finally, the vector $\textbf{v}$ is fed into an MLP classifier to determine the sentiment polarity of the input sentence.

For the evaluation of Tree-LSTM, we directly use the constituency Tree-LSTM model in \cite{tai2015improved} which has the same structure with that described in Section \ref{desc:treelstm}. Particularly, the Tree-LSTM models in \cite{tai2015improved} only use the state of root node as the final sentence representaion.

\section{Experiments on NLI}\label{test_nli}
%In this section, we first compare the performance of our proposed SNELS model with LSTM-RNNs and Tree-LSTMs for the sentence encoding of the  SNLI task.
%since SNELS is a model with 2-layer hierarchical chain structure and each layer can be analogic to LSTM in some way.
%Then we compare the SNELS with other published state-of-art results. Finally we analyzed the 'chunking' patter of SNELS model on NLI task to decide wether our SNELS can utilized some regular latent structure information.
\subsection{Experimental Setup}
In our experiments, the Stanford Natural Language Inference (SNLI) dataset \cite{bowman2015large} was adopted.
Following previous work \cite{bowman2015large}, the sentence pairs lacking consensus among multiple human annotators were removed, and this dataset was split into a training set, a development set, and a test set with $549,367$, $9,842$, and $9,824$ sentence pairs, respectively.

The entire NLI model was constructed following the introduction in Section \ref{nli_descripttion}.
Various sentence encoders, including LSTM-RNN, BLSTM-RNN, Tree-LSTM, and the proposed SNELSD model, were integrated to achieve the sentence encoding in (25) and (26).
The constituency parse trees for constructing Tree-LSTM models were produced using the Stanford PCFG Parser 3.5.3 \cite{klein2003accurate}.
All word embeddings and the hidden state vectors of sentence encoders had 300 dimensions. Specifically, the hidden states of the BLSTM-RNN model had 600 dimensions since it was composed of two unidirectional LSTM-RNNs along different directions.
The model parameters of the sentence encoders were estimated in an end-to-end manner together with the parameters of other parts of the NLI model.
The word embeddings were initialized by pre-trained \emph{300-D Glove 840B} vectors \cite{pennington2014glove}.
Out-of-vocabulary (OOV) words were randomly initialized with Gaussian-distributed samples.
Cross-entropy was adopted as the loss function for model training, and the Adam \cite{kingma2014adam} method was adopted for optimization.
The first momentum was set to be 0.9 and the second to be 0.999 in Adam optimization.
The initial learning rate was 0.0004, and the batch size was 128.
Dropout with a rate of 0.5 was applied to the MLP layer and the word embedding layer.
%ALL hidden states of sentence encoding models and BiLSTM, and word embeddings are 300 dimensions.

\begin{table}[!t]
\renewcommand{\arraystretch}{1.3}
\caption{Accuracies (\%) of using different sentence encoders on SNLI.}
\label{compare_nli}
\centering
\begin{tabular}{p{0.8cm}p{4.5cm}cc}
\hline
 & Sentence Encoder & Train & Test\\
\hline
\multicolumn{1}{c}{\multirow{6}{*}{\emph{A}}} & 1-layer LSTM & 91.6 & 87.7\\
 & 1-layer BLSTM & 92.5 & 87.7\\
 \cline{2-4}
 & 2-layer LSTM & 91.3 & 86.9\\
 & 2-layer BLSTM & 91.6 & 87.5\\
 \cline{2-4}
 & Tree-LSTM & 92.0 & 87.3\\
 \cline{2-4}
 & SNELSD & 93.3 & 87.3\\
\hline
\multicolumn{1}{c}{\multirow{5}{*}{\emph{B}}} & [1-layer LSTM; word embedding] & 92.9 & 87.5\\
 & [1-layer BLSTM; word embedding] & 92.8 & 87.7\\
 \cline{2-4}
 & [2-layer LSTM; word embedding] & 93.9 & 87.4\\
 & [2-layer BLSTM; word embedding] & 92.8 & 87.7\\
 \cline{2-4}
 & [SNELSD; word embedding] & 92.2 & \textbf{88.0}\\
\hline
\multicolumn{1}{c}{\multirow{2}{*}{\emph{C}}} & [2-layer LSTM; 1-layer BLSTM] & 95.1 & 87.8\\
\cline{2-4}
 & [SNELSD; 1-layer BLSTM] & 94.5 & \textbf{88.0}\\
\hline
\end{tabular}
\end{table}

\subsection{Performance of SNELSD in Stand-Alone Mode}\label{nli_alone}
We compared the performance of using SNELSD in stand-alone mode for sentence modeling with that of using the LSTM-RNN and Tree-LSTM models.
Considering that an SNELSD has a two-layer sequential structure, four different LSTM-RNNs, including  1-layer and 2-layer unidirectional LSTM-RNNs and 1-layer and 2-layer BLSTM-RNNs, were used for comparison.
%In this experiment, we compared the SNELS with LSTMs and Tree-LSTM models on SA task since our SNELS model is a sequence-based model like LSMTs and also has a hierarchical structure like Tree-LSTM in some way. For comparison, we only replace the SNELS with other models for sentence encoding.
The NLI accuracies of using these sentence encoders are shown in Part A of Table \ref{compare_nli}.

First, we observe that the SNELSD model achieved similar performance with the Tree-LSTM model, which exploited syntactic information explicitly.
The SNELSD model performed better than the 2-layer LSTM-RNN, which also had a 2-layer unidirectional chain structure.
%Since both models had a 2-layer unidirectional chain structure, this result shows the potential advantage of SNELSD over LSTM-RNN.
However, the accuracy of the SNELSD model was still lower than that of the 2-layer BLSTM-RNN. One possible reason for this result is the lack of backward processing in SNELSD.
%we can find that the Tree-LSTM which exploits the syntactic information do not outperform SNELS. Also, the results show that , which indicates that our SNELS can capture some information that 2-layer LSTM cannot. For the reason why 2-layer BiLSTM behaves better than SNELS, we think that the BiLSTM can take advantage of the backward information in sentences but our SNELS cannot.

Furthermore, it can be found that both 1-layer LSTM-RNN and BLSTM-RNN outperformed their 2-layer counterparts and the two models with structured descriptions (i.e., Tree-LSTM and SNELSD).
Although it has been observed empirically that multi-layer RNNs achieved better performance than 1-layer RNNs on some tasks, such as machine translation in an encoder-decoder framework \cite{sutskever2014sequence}, the gains of using stacked RNNs for sentence modeling are still not theoretically clear \cite{DBLP:journals/jair/Goldberg16}.
One possible reason for the superiority of 1-layer RNNs in our experiment  is that stacking RNN layers or introducing a structured description leads to abstract and compositional representations of sentences, which may be inappropriate for the NLI task.
Some less compositional descriptions, such as the meanings of individual key words in sentences, are also important for deciding the inference relationship between two sentences.

In terms of the training efficiency of different models, we observed that the training speed of SNELSD was similar with 2-layer LSTM, but significantly faster than Tree-LSTM in our experiments.
As discussed in Section 1, the sequential architecture of SNELSD  guarantees the efficiency of model training.
%Therefore,  the performance of using SNELSDs in joint mode was further investigated by experiments, which will be introduced in next subsection.

%we think that the sentence representations encoded by 2-layer LSTM/BiLSTM are more abstract and condensed than that encoded by 1-layer LSTM/BiLSTM, in other words, the outputs of 2-layer LSTM/BiLSTM are higher-level feature, and maybe the NLI task also mainly depends on low-level feature since that when we decide the inference relationship of two sentences we will match the information from some key words in sentences.

\subsection{Performance of SNELSD in Joint Mode}\label{nli_joint}
The performance of using SNELSD in joint mode was further investigated through experiments.
Here, the hidden state vectors produced by the SNELSD model were concatenated with unprocessed word embeddings or with the output of a 1-layer BLSTM-RNN to achieve the sentence encoding in (25) and (26).
Some models that combined the output of LSTM-RNNs, BLSTM-RNNs, and word embeddings  for sentence modeling were also constructed for comparison.
The results of these models are shown in Parts B and C of Table \ref{compare_nli}.

%As discussed above, we know that all 2-layer structured SNELS/LSTM/BiLSTM may suffer from lack of low-level information from sentences, so here we try to concatenate the sentence representation with the original word embedding sequence and use the results of this concatenation as the final sentence representation. We test this method on SNLI dataset, and the results are shown in Part B of Table \ref{compare_nli}. Further more, we concatenated outputs of two encoding models to find that whether 1-layer model and 2-layer model can compensate each other, and results are shown in Part C of Table \ref{compare_nli}.

Comparing the results in Part B of Table \ref{compare_nli} with those in Part A,
we find that the concatenation with word embeddings failed to improve the performance of 1-layer RNNs, whereas it increased the accuracy of 2-layer RNNs and the SNELSD model.
Comparing the results in Part C with those in Part A, we can observe the positive effects of concatenating the output of 1-layer BLSTM-RNN with 2-layer LSTM-RNN and the SNELSD model.
These results demonstrate the importance of utilizing a comprehensive sentence representation for the NLI task.

As shown in Parts B and C of Table \ref{compare_nli}, the proposed joint-mode SNELSD model achieved the highest accuracy of $88.0\%$ among all evaluated sentence encoders.
This result demonstrates that by introducing a latent structured description, the SNELSD model is able to provide some useful structured semantic information for  NLI  that conventional sequential sentence encoders may ignore.

%We can find that all 2-layer structured SNELS/LSTM/BiLSTM after concatenated with word embeddings have an improved performance on SNLI especially for the 2-layer LSTM and SNELS models. And SNELS achieved the best performance than other LSTMs works. However, the 1-layer LSTM and BiLSTM after concatenated with word embeddings do not behaves better after joint with word embeddings. This implies that higher-level information is really important for NLI task. From results shown in Part C of Table \ref{compare_nli}, we can find that the performance of 1-layer BiLSTM after concatenated with SNELS has a more significant improvement than that concatenated with 2-layer LSTM, which means that features from different levels can compensate each other on NLI task.

\subsection{Comparison with State-of-the-Art Results}
\begin{table}[!t]
\renewcommand{\arraystretch}{1.3}
\caption{Accuracies (\%) of using different models on SNLI.}
\label{compare2}
\centering
\begin{tabular}{p{5.7cm} c c}
\hline
Models & Train & Test\\
\hline
(1) decomposable attention model \cite{parikh2016decomposable} & 89.5 & 86.3\\
(2) NTI-SLSTM-LSTM \cite{munkhdalai2016neural} & 88.5 & 87.3\\
(3) EBIM \cite{chen2016enhancing} & 92.9 & 87.7\\
(4) ESIM \cite{chen2017enhancing} & 92.6 & 88.0\\
(5) BiMPM \cite{wang2017bilateral} & -- & 86.9\\
{(6) [SNELSD; 1-layer BLSTM]} & 92.2 & 88.0 \\
{(7) [SNELSD; 1-layer BLSTM]*} & 94.1 & \textbf{88.3} \\
\hline
(8) ESIM + Syntactic tree-LSTM (Ensemble)\cite{chen2017enhancing} & 93.5 & 88.6 \\
(9) BiMPM (Ensemble) \cite{wang2017bilateral} & -- & \textbf{88.8}\\
{(10) [SNELSD; 1-layer BLSTM]* (Ensemble)} & 93.3 & 88.7 \\
\hline
\end{tabular}
\end{table}

\begin{figure*}[!t]
\centering
\includegraphics[width=5.5in]{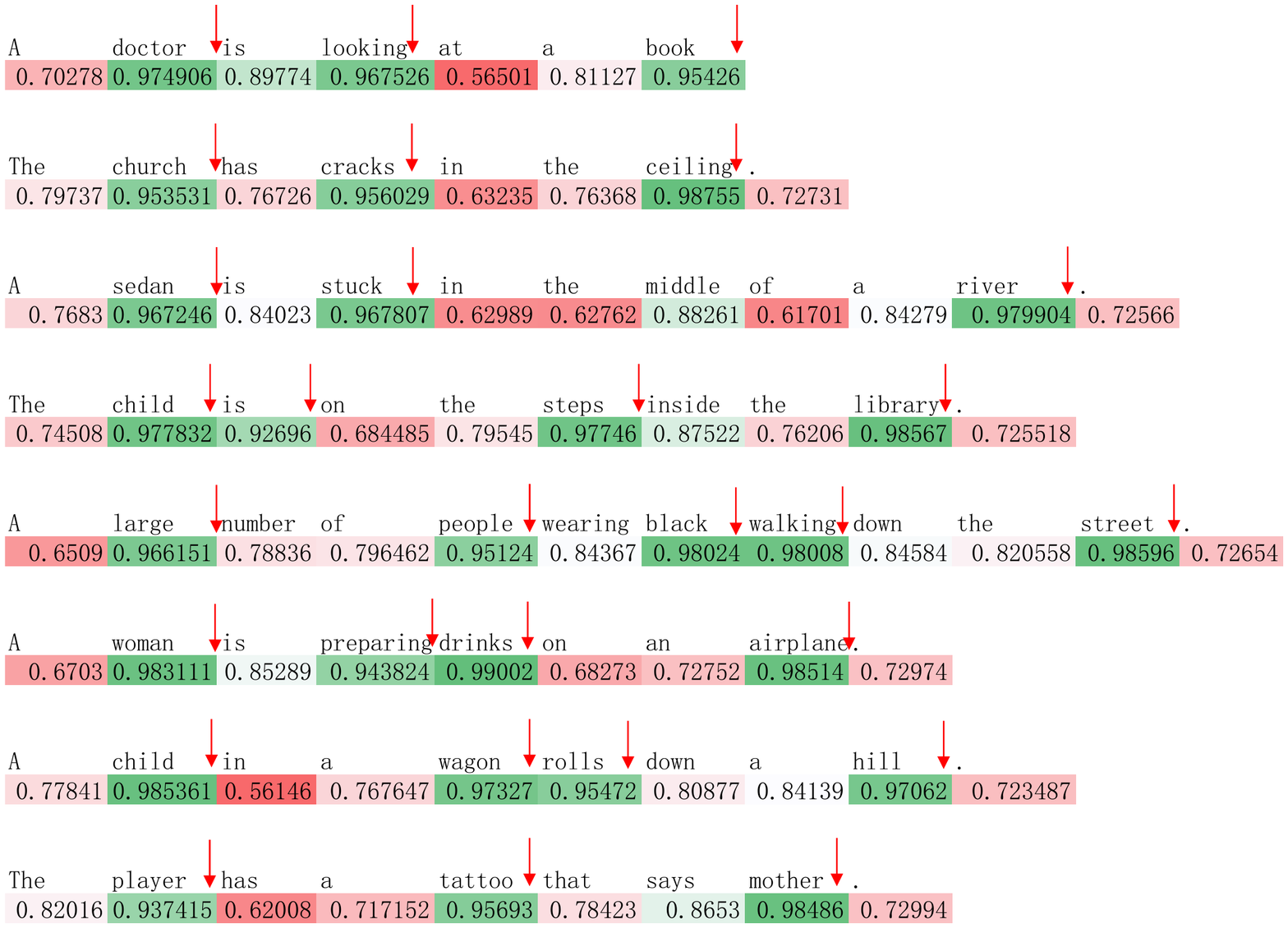}
\caption{
Visualization of latent word chunks in SNELSD for NLI. Each line corresponds to a hypothesis sentence in the SNLI test set.
The value of the chunk boundary indicator $r_t$ is written below the $t$-th word in the sentence together within a red-green scale color block.
When the color block is more red, the value of $r_t$ is smaller, which means that it is more unlikely to have a word chunk boundary after the word $\textbf{x}_t$.
In contrast, when the color block is more green, the value of $r_t$ is larger, and the probability of  a chunk boundary existing after the word $\textbf{x}_t$ is higher.
For better illustration, red arrows are added to indicate the positions where $r_t$ is higher than $0.9$.
%Visualization of 'chunking' pattern of SNELS model on NLI task. Sentences shown in this figure are hypothesis sentences in the test dataset of Stanford Natural Language Inference. The value of boundary indicate coefficients $r_t$ are written below the corresponding words together with a red-green scale color block. With the color block more red, the vaule of $r_t$ is more smaller, so there tends not to exist the boundary at the right position of word $\textbf{x}_t$. Contrarily, with the color block more green, the value of $r_t$ is more bigger, the probability of the existence of boundary at the right position of word $\textbf{x}_t$ is more greater.
}
\label{fig:r_nli}
\end{figure*}
Here, the proposed method using SNELSD models was compared with other published state-of-the-art methods on the benchmark SNLI task.
The results are shown in Table \ref{compare2}. %For the sake of fairness
All the models compared here utilized the framework of word-by-word attention between the semantic representation of two sentences. %, which we call \emph{soft alignment} in this paper.
The methods marked with (1)-(7) adopted a single model for NLI, and the methods marked with (8)-(10) ensembled multi-models for decision making.

The decomposable attention model \cite{parikh2016decomposable} marked with (1) can be treated as a simplified edition of EBIM.
The most obvious difference between them is that the decomposable attention model employs feedforward neural networks, %does not use any sequential model but feed forward neural networks,
whereas the EBIM model uses BLSTM-RNNs to encode the sentences and to compose the inference information.
We can find that this model behaved considerably worse than the other models shown in Table \ref{compare2}.

The NTI-SLSTM-LSTM model \cite{munkhdalai2016neural} marked with (2)  utilized a full binary tree structure, which incorporates the sequential LSTM-based encoding, recursive networks and complicated combination of attention.
However, this model still performed worse than EBIM and joint-mode SNELSD. The tree structure also made training of the model time consuming.

The EBIM model \cite{chen2016enhancing} marked with (3)  is the same as the model using 1-layer BLSTM-RNN as the sentence encoder in Table \ref{compare_nli}.
The ESIM model \cite{chen2017enhancing} marked with (4)  is an improved version of the EBIM model \cite{chen2016enhancing}. %(work \cite{chen2017enhancing} is an updated edition of \cite{chen2016enhancing}),
The only difference is that the ESIM  model introduces a 1-layer feed-forward neural network with the ReLU activation function to reduce the dimensions of $\textbf{m}_a$ and $\textbf{m}_b$ calculated at the inference collection step described in Section \ref{nli_descripttion}.
%are reported by the original author who proposed the EBIM\cite{chen2016enhancing}, we can find that the performance of EBIM reported in chen et al.(2016)\cite{chen2016enhancing} is almost the same with the 1-layer BiLSTM model we realized in Table \ref{compare_nli}.

The BiMPM model \cite{wang2017bilateral} marked with (5)  is almost the same as EBIM.
There are two main differences between these models.
The first difference is that the original word representation used in BiMPM is the concatenation of pre-trained GloVe word embeddings and character-composed embeddings derived from an LSTM-RNN \cite{wang2017bilateral}.
The second difference is that BiMPM adopts a more complicated matching operation than EBIM. %, which corresponds to \emph{soft alignment} and \emph{subcomponent inference collection} operations in EBIM.
We find that both EBIM and joint-mode SNELSD were superior to BiMPM.

The models marked with (6) and (7) utilized joint-mode SNELSD in an EBIM and ESIM fashion, respectively.
When combining joint-mode SNELSD with ESIM, the outputs of SNELSD and 1-layer BLSTM-RNN were concatenated after inference collection and dimension reduction.
It can be observed that models (6) and (7) achieved higher NLI accuracies than their counterparts using 1-layer BLSTM-RNN for sentence modeling.
%The model marked with (6) has been reported in Table \ref{compare_nli}, we can find it achieves the same performance with ESIM. For model marked with (7), we just adopted the dimension reduction used in ESIM %\cite{chen2017enhancing} on model marked with (6), specifically, the feature concatenation of SNELSD with 1-layer BLSTM is not conducted after sentence encoding but after inference collection and dimension reduction here.
Furthermore, combining joint-mode SNELSD in an ESIM fashion achieved an accuracy of $88.3\%$, which is the best result among all single-model methods listed in Table \ref{compare2}.
A close examination on the test set errors made by models (4) and (7) shows that about 3.5\% test set samples were misclassfied by model (4) while classified correctly by model (7).
On the contrary, about 3.2\% test set samples were misclassfied by model (7) while classified correctly by model (4). More than 70\% of the errors made by these two models were the same.
This implies that there still exist common deficiencies with these two models.
We also observed that some of the common errors made by these two models were due to the lack of background and commonsense knowledge during the inference.
How to integrate external knowledge represented by lexical databases, such as WordNet \cite{miller1995wordnet}, into the sentence modeling and matching of NLI is a topic worth further investigation.

The model marked with (8)  assembled an original ESIM and a modified ESIM using a syntactic Tree-LSTM for sentence encoding and information composition \cite{chen2017enhancing}.
It achieved a significant improvement over the single ESIM model.
The model marked with (9) assembled two BiMPMs with identical architectures \cite{wang2017bilateral}.
Similarly, we constructed an ensemble model using two instances of  model (7) trained independently.
Thus, two posterior probability vectors corresponding to the three options of NLI were returned for each sentence.
Similar to the model ensemble strategy used in \cite{chen2017enhancing,wang2017bilateral}, these two posterior probability vectors were averaged to obtain the final one for decision.
It can be observed that this model ensemble achieved an accuracy of $88.7\%$, which was higher than the accuracy of $88.3\%$ using a single model, and a little bit higher than ensembled ESIM utilizing syntactic tree-LSTM\cite{chen2017enhancing}.
Although the differences between the models marked with (8) and (10) is insignificant, %the model marked with (8) needs a syntactic parser.
the SNELSD model marked with (10)  has other benefits, such as no dependency on extra syntactic parsers and simple end-to-end training framework.
%Additionally, it can be found that our ensembled SNELSD model didn't achieved so much improvement like model (4) and (5), we think that it's maybe because the training of our single SNELSD model is more stable and better in performance, since it achieved the best prediction accuracy among all single models.
Additionally, it can be observed that our SNELSD model didn't achieve as much gain as the BiMPM model from the ensemble.
One possible reason is that the performance of our single SNELSD model was quite stable among its instances trained independently.

\subsection{Visualization of Latent Word Chunks in SNELSD for NLI}
As described in Section \ref{intro}, an SNELSD model segments sentences into word chunks in a task-dependent manner for sentence modeling.
The word chunks are latent and described by a sequence of stochastic chunk boundary indicators.
%we know that the 'chunking' in SNELS is a end-to-end manner without any syntactic information and we think the structure information captured by SNELS is latent in sentence and task-dependent. So
Here, the calculated boundary indicators $r_t$ for several sentence examples are displayed to visualize the latent word chunks considered in SNELSD modeling. %chunking pattern of SNELS, specially, it is the boundary indicator vetor $\textbf{r}$.
The single SNELSD model in Part A of Table \ref{compare_nli} was adopted here, and the results are shown in Fig. \ref{fig:r_nli}.
In this figure, each line corresponds to a hypothesis sentence in the SNLI test set.
The value of $r_t$ is written below the $t$-th word in the sentence together within a red-green scale color block.
When the color block is more red, the value of $r_t$ is smaller, which means that it is more unlikely to have a word chunk boundary after the word $\textbf{x}_t$.
Conversely, when the color block is more green, the value of $r_t$ is larger, and the probability of  a chunk boundary existing after the word $\textbf{x}_t$ is higher.
For better illustration, red arrows are added in Fig. \ref{fig:r_nli} to indicate the positions where $r_t$ is higher than $0.9$.

As shown in Fig. \ref{fig:r_nli}, the chunking patterns provided by the SNELSD model can partially capture the main grammar structure of a sentence.
Most of the phrase boundaries have a high value of $r_t$, which means that they are also judged to have high possibilities of being word chunk boundaries in the SNELSD model.
This result is reasonable since the SNELSD model learns how to segment sentences into word chunks by task-dependent and end-to-end model training,
and the syntax-related information should be useful for the NLI task.

%As shown in Fig. \ref{fig:r_nli}, it's the blocking pattern of part of hypothesis sentences on NLI task. We can find that the chunking pattern do not adhere rigidly to syntactic structure, though consistent to it in some way, but more obey the rule that blocking sentences into parts correlated with characters/subjects, events/actions and sites/objects. And this rule seems to be the key point when we decide the inference relationship between two difference sentences, because the comparison and matching between two sentences are just conducted on that three basic parts (characters/subjects, events/actions and sites/objects).
\begin{table}[!t]
\renewcommand{\arraystretch}{1.3}
\caption{%Comparative performance (accuracy) with standard deviation on SST dataset of SNELS model with LSTMs and Tree-LSTM models, all models are trained and tested on sentence-level fine-grained labels.
Average accuracies (\%) with standard deviations of using different sentence encoders on SA.}
\label{compare_sa}
\centering
\begin{tabular}{p{0.8cm}p{4.5cm}c}
\hline
 & Sentence Encoder & Test(std)\\
\hline
\multicolumn{1}{c}{\multirow{5}{*}{\emph{A}}} & 1-layer LSTM  & 46.7(0.89)\\
 & 1-layer BLSTM &  46.9(1.30)\\
 \cline{2-3}
 & 2-layer LSTM & 46.2(1.32)\\
 & 2-layer BLSTM & 46.6(1.80)\\
 \cline{2-3}
 & Tree-LSTM \cite{tai2015improved} & {47.7}(0.67)\\
 \cline{2-3}
 & SNELSD & 46.8(1.28)\\
\hline
\multicolumn{1}{c}{\multirow{5}{*}{\emph{B}}} & [1-layer LSTM; word embedding] & 46.2(1.27)\\
 & [1-layer BLSTM; word embedding]  & 46.8(1.46)\\
 \cline{2-3}
 & [2-layer LSTM; word embedding]  & 46.0(1.00)\\
 & [2-layer BLSTM; word embedding] & 46.1(1.35)\\
 \cline{2-3}
 & [SNELSD; word embedding] & 46.7(0.90)\\
\hline
\multicolumn{1}{c}{\multirow{2}{*}{\emph{C}}} & [2-layer LSTM; 1-layer BLSTM] & 47.0(1.14)\\
 & [SNELSD; 1-layer BLSTM] & \textbf{47.9}(1.24)\\
\hline
\end{tabular}
\end{table}
\section{Experiments on SA}\label{test_sa}
In contrast to NLI, which concerns the relationship between two sentences,
sentiment analysis (SA) is to classify a single sentence.
In this section, %in order to evaluate our DLCE model's ability to sentence modelling more comprehensively and objectively,
we performed experiments to compare the SNELSD, LSTM-RNN and Tree-LSTM models on SA similar to Section \ref{test_nli}.

\subsection{Experimental Setup}
We used the Stanford Sentiment Treebank (SST) dataset and the same data split as in Socher et al. (2013) \cite{socher2013recursive},
in which the number of training, development and test sentences were $8,544$, $1,101$, and $2,210$, respectively.
%It should be noted that we did not use the phrase-level labels to train our model, since out SNELS model will learn to split the sentence into word blocks.
Since the focus of this paper is on sentence modeling, we excluded the phrase-level samples in the SST dataset and only used sentence-level samples for evaluating different sentence encoders.
%So the phrases in the parse trees of 8,544 sentences in training dataset will not be included in our training process, though many other previous work used these phrases-level labels.

The framework introduced in Section IV-B was followed to construct the SA model.
Different sentence encoders were used to fulfill the sentence encoding in (37) for comparison.
During model training, cross entropy was chosen as the loss function for optimization, and the Adadelta \cite{zeiler2012adadelta} method was used for the optimization process.
The $\epsilon$ was set to $1e-6$ and $\rho$ was set to $0.95$ for Adadelta optimization.
%The initial learning rate was 0.05 and the batch size was 16.
The batch size was 16.
%ALL hidden states of sentence encoding models and word embeddings are 300 dimensions.
All word embeddings and the hidden state vectors of sentence encoders had 300 dimensions.
Specifically, the hidden states of the BLSTM-RNN model had 600 dimensions since it was composed of two unidirectional LSTM-RNNs along different directions.
The drop-out strategy was not applied %because of the insufficient training data and
to avoid the instability among different training trials.
%Specifically, we use the Adam \cite{kingma2014adam} method was for optimization of Tree-LSTM model. The first momentum was set to be 0.9, the second to be 0.999 and the initial learning rate was set to be 0.0004 in Adam optimization.
The word embeddings were initialized by pre-trained \emph{300D GloVe 840B} vectors \cite{pennington2014glove}.
The results are shown in Table \ref{compare_sa}, where the evaluated sentence encoders were the same as those shown in Table \ref{compare_nli}.
All the results in Table \ref{compare_sa} are the averages and standard deviations of 40 training trials on the test set.
%because the training results are instable, so we train each model 40 times and average the test results.
\begin{table}[!t]
\renewcommand{\arraystretch}{1.3}
\caption{Accuracies (\%) of using different models on SA.}
\label{compare_sa2}
\centering
\begin{tabular}{p{5.7cm} c c}
\hline
Models & Test\\
\hline
(1) S-LSTM \cite{zhu2015long} & 43.5\\
(2) CNN-Word2Vec \cite{zhang2015sensitivity} & 47.1\\
(3) CNN-Glove \cite{zhang2015sensitivity} & 45.7\\
\hline
(4) SNELSD  & 46.8\\
(5) [SNELSD; 1-layer BLSTM]  & \textbf{47.9}\\
\hline
\end{tabular}
\end{table}

\begin{figure*}[!t]
\centering
\includegraphics[width=5.8in]{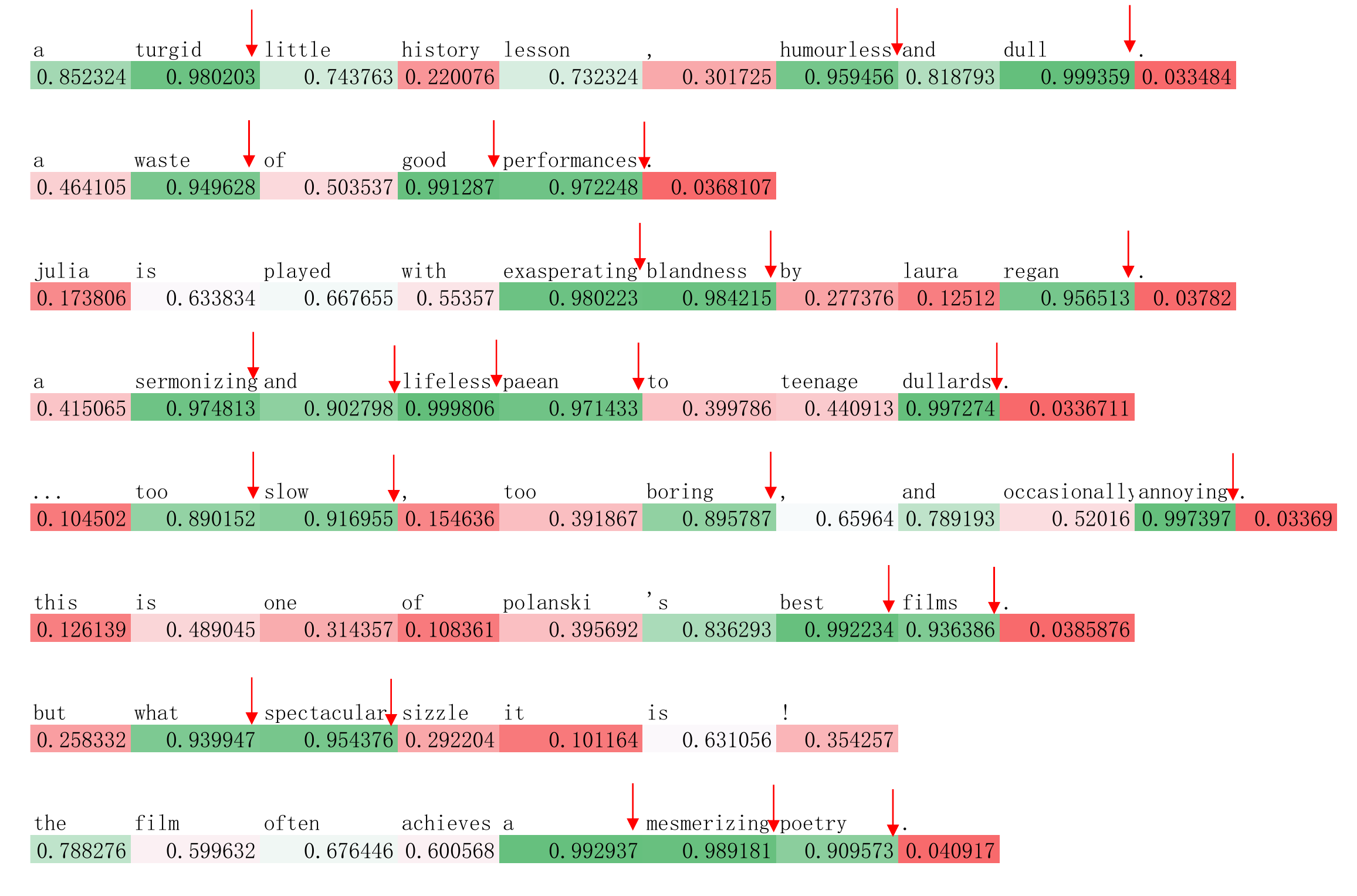}
\caption{
%Visualization of 'chunking' pattern of SNELS model on SA task. Sentences shown in this figure are included in the test dataset of Stanford Sentiment Treebank. The value of boundary indicate coefficients $r_t$ are written below the corresponding words together with a red-green scale color block. With the color block more red, the vaule of $r_t$ is more smaller, so there tends not to exist the boundary at the right position of word $\textbf{x}_t$. Contrarily, with the color block more green, the value of $r_t$ is more bigger, the probability of the existence of boundary at the right position of word $\textbf{x}_t$ is more greater.
Visualization of latent word chunks in SNELSD for SA. Each line corresponds to a hypothesis sentence in the SST test set.
The value of the chunk boundary indicator $r_t$ is written below the $t$-th word in the sentence together within a red-green scale color block.
When the color block is more red, the value of $r_t$ is smaller, which means that it is more unlikely to have a word chunk boundary after the word $\textbf{x}_t$.
Conversely, when the color block is more green, the value of $r_t$ is larger, and the probability of  a chunk boundary existing after the word $\textbf{x}_t$ is higher.
For better illustration, red arrows are added to indicate the positions where $r_t$ is higher than $0.9$.}
\label{fig:r_sa}
\end{figure*}
\subsection{Performance of SNELSD in Stand-Alone Mode}
We compared the performance of using SNELSD in stand-alone mode for the sentence encoding in SA with that of using the LSTM-RNN and Tree-LSTM models.
The results are shown in Part A of Table \ref{compare_sa}.
%In this experiment, we compared the SNELS with LSTMs and Tree-LSTM models on SA task just like the settings of experiments in Section \ref{test_nli}. For comparison, we just replace the SNELS with LSTMs or Tree-LSMT for sentence encoding. The results are shown in part A of Table \ref{compare_sa}.

It can be observed that the Tree-LSTM model achieved the best performance among all evaluated sentence encoders.\footnote{This result was derived using the codes for training the constituency Tree-LSTM model in \cite{tai2015improved}.}
Similar to the NLI results shown in Table \ref{compare_sa}, the 2-layer LSTM and BLSTM models also performed worse  than their 1-layer counterparts on the SA task, which is consistent with the experimental results in \cite{tai2015improved}.
However, our SNELS model, which had a similar 2-layer unidirectional chain structure as the 2-layer LSTM-RNN, outperformed both 2-layer LSTM and 2-layer BLSTM
and obtained almost the same test accuracy as 1-layer LSTM and BiLSTM.
This result implies that the detection layer and description layer in our SNELS model may be capable of utilizing latent structure information in sentences that conventional LSTM units cannot capture.

\subsection{Performance of SNELSD in Joint Mode}
Similar to Section \ref{nli_joint}, we further evaluated the performance of the SNELSD model working in joint mode, and the results are summarized in Parts B and C of Table  \ref{compare_sa}.
%As discussed above, we know that 2-layer LSTM/BiLSTM has a weaker performance than 1-layer LSTM/BiLSTM. Similar to experiments in Section \ref{test_nli}, we conduct the experiments that concatenating sentence representation with original word embedding to determine whether the our SNELS model needs the low-level information on sentiment analysis (SA) task as well as on NLI task. Results are shown in Part B of Table \ref{compare_sa}.

Comparing the results in Part B of Table \ref{compare_nli} with those in Part A,
it can be found that the concatenation with word embeddings degraded the performance of all sentence encoders,
particularly for the 1-layer LSTM and 2-layer BLSTM models, which had a significant  accuracy reduction of approximately $0.5\%$.
This result is inconsistent with the results on the NLI task and
implies that to simply consider the representation of all individual words in the sentence modeling may be inappropriate for the SA task.
%As for SNELS, it seems more stable than other models with only $0.1\%$ reduce.

Examining the results in Part C of Table \ref{compare_sa},
it can be observed that both 2-layer LSTM and our proposed SNELSD model benefited from the concatenation with another 1-layer BLSTM encoder.
The joint-mode SNELSD model achieved an accuracy of $47.9\%$, which was the highest accuracy among all the evaluated sentence encoders.
%Also, we conducted the experiments shown in Part C of Table \ref{compare_sa}, it's concatenation of 2-layer's representation and 1-layer's representation, we can find that 1-layer BiLSTM after concatenated with soft-SNELS has a significant improved performance with $1.0\%$ uplift, contrarily, that concatenated with 2-layer LSTM does not, which
This result further confirms that the SNELSD model can truly capture some information that ordinary LSTMs models cannot capture.

\subsection{Comparison with Published Results}
Most existing works on the Stanford Sentiment Treebank (SST) dataset trained their models using both phrase-level and sentence-level labels.
The best published results we can found that only adopted sentence-level samples of this dataset for model training and testing are compared in Table \ref{compare_sa2}.

The S-LSTM \cite{zhu2015long} marked with (1) is also a tree-structured LSTM model, which is similar to our Tree-LSTM model but initialized their word embeddings randomly.
%It can be found that S-LSTM achieves  similar results with the Tree-LSTM shown in Part A of Table \ref{compare_sa}.
The models marked with (2) and (3) are CNN-based models \cite{zhang2015sensitivity} using Word2Vec \cite{mikolov2013distributed} and GloVe \cite{pennington2014glove} to initialize word embeddings, respectively.
%We can find that the Non-static word2vec-CNN \cite{zhang2015sensitivity} outperforms SNELSD but the joint-mode SNELSD still achieves the best performance among these models.
We can find that the joint-mode SNELSD model achieved the best performance among all these models.

\subsection{Visualization of Latent Word Chunks in SNELSD for SA}
We also visualized the chunking pattern of applying the SNELSD model on the SA task.
The values of $r_t$ of the SNELSD model in Part A of Table \ref{compare_sa} were used for illustration, and the results are shown in Fig. \ref{fig:r_sa}.
The first 5 sentences had negative sentiment labels, whereas the last 3 sentences were positive.
%We use the same auxiliary method to point out boundary positions with biggest probability.
From this figure, it can be found that the word chunk boundaries tend to exist following the words with strong emotion polarities.
%This is somewhat consistent with how human make decisions for sentiment analysis,
%that when we read to a word indicates negative or positive feeling we tend to first pause and then continue the reading process.
This result is reasonable since more attention should be given to these words for sentiment analysis.

Comparing Fig. \ref{fig:r_sa} with Fig. \ref{fig:r_nli}, we can see that the chunking pattern provided by our SNELSD model is truly task dependent, which illustrates the flexibility of our SNELSD model to be compatible with different tasks. % with very different intrinsic property.

%\section{Some further analysis on SNELS}
%\subsection{Is the 'chunking' in SNELS really task-dependent ?}
%From comparison analysis experiments on SA and NLI task, we know that there are inconsistent results between sentiment analysis and natural language inference, which indicates that there exist task-dependent aspects when we cope with different tasks using the same model. Moreover, our SNELS model learns how to chunk sentences totally by an end-end manner without utilizing any outer syntactic information and constraint, so we think the chunking of SNELS is task-dependent not like the chuning task in NLP, which is syntax-dependent. And from the visualization results of 'chunking' pattern in Section \ref{test_nli} and \ref{test_sa}, we can know that the SNELS model can truly capture some regular structure pattern of sentences on both NLI and SA task.
%
%Here, in order to verify our guess about the task-dependent 'chunking' of SNELS more objectively, we manually created two sentences and processed these two sentences using model marked with (6) in Table \ref{compare2} which was trained on NLI and SNELS model in Part A of Table \ref{compare_sa} which was trained on SA task. Then we visualize the 'chunking' pattern on these two sentences of both models.
%%\begin{figure*}[!t]

\section{Conclusion}
This paper has proposed a novel sentence encoding model named sequential neural encoder with latent structure description (SNELSD).
This model has a 2-layer hierarchical chain structure and splits sentences into latent word chunks by end-to-end learning. %And on practical tasks, here
A natural language inference (NLI) task and a sentiment analysis (SA) task are introduced to evaluate the proposed SNELSD model.
The experimental results show that the proposed SNELSD model can fit these tasks very well and that joint-mode SNELSDs outperform ordinary LSTM models by capturing and utilizing the latent structured information of sentences in a task-dependent manner.
Applying the proposed SNELSD model to more tasks, such as paragraph comprehension and question answering, will be the tasks of our future work.
\ifCLASSOPTIONcaptionsoff
  \newpage
\fi

% trigger a \newpage just before the given reference
% number - used to balance the columns on the last page
% adjust value as needed - may need to be readjusted if
% the document is modified later
%\IEEEtriggeratref{8}
% The "triggered" command can be changed if desired:
%\IEEEtriggercmd{\enlargethispage{-5in}}

% references section

% can use a bibliography generated by BibTeX as a .bbl file
% BibTeX documentation can be easily obtained at:
% http://mirror.ctan.org/biblio/bibtex/contrib/doc/
% The IEEEtran BibTeX style support page is at:
% http://www.michaelshell.org/tex/ieeetran/bibtex/
\bibliographystyle{IEEEtran}
 %argument is your BibTeX string definitions and bibliography database(s)
\bibliography{IEEEabrv,reference}
%
% <OR> manually copy in the resultant .bbl file
% set second argument of \begin to the number of references
% (used to reserve space for the reference number labels box)
%\begin{thebibliography}{1}
%
%\bibitem{IEEEhowto:kopka}
%H.~Kopka and P.~W. Daly, \emph{A Guide to \LaTeX}, 3rd~ed.\hskip 1em plus
%  0.5em minus 0.4em\relax Harlow, England: Addison-Wesley, 1999.
%
%\end{thebibliography}

% biography section
%
% If you have an EPS/PDF photo (graphicx package needed) extra braces are
% needed around the contents of the optional argument to biography to prevent
% the LaTeX parser from getting confused when it sees the complicated
% \includegraphics command within an optional argument. (You could create
% your own custom macro containing the \includegraphics command to make things
% simpler here.)
%\begin{IEEEbiography}[{\includegraphics[width=1in,height=1.25in,clip,keepaspectratio]{mshell}}]{Michael Shell}
% or if you just want to reserve a space for a photo:
\begin{IEEEbiography}[{\includegraphics[width=1in,height=1.25in,clip,keepaspectratio]{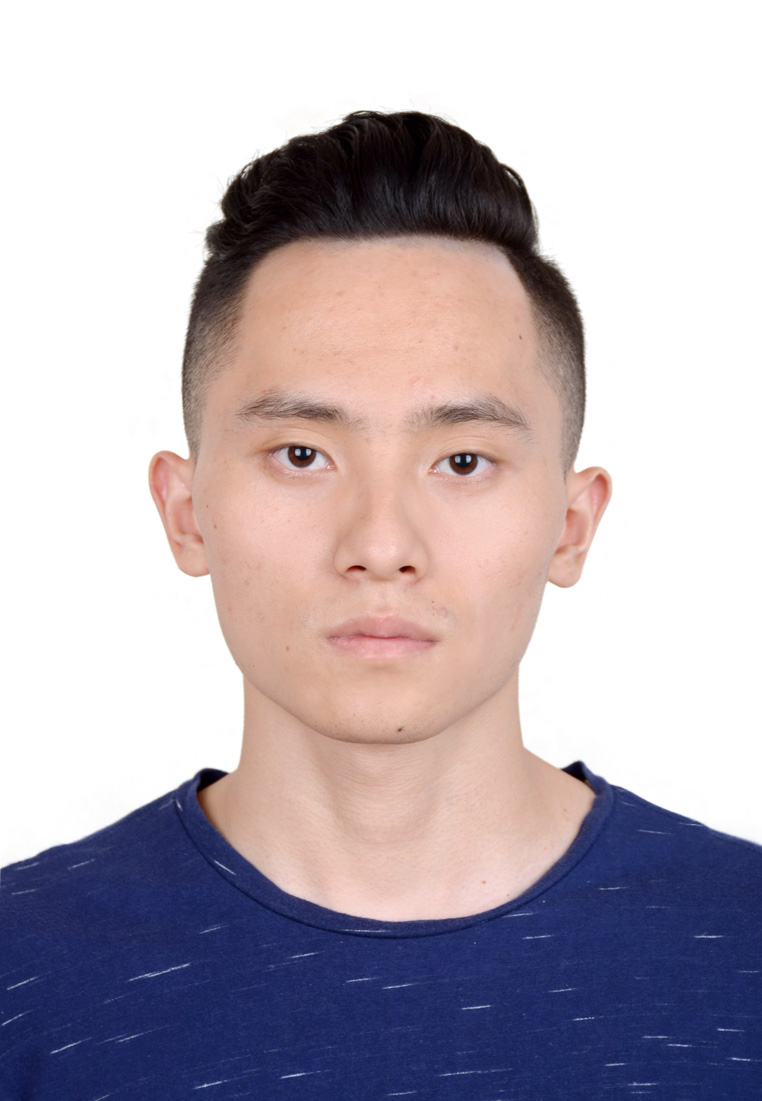}}]{Yu-Ping Ruan} received the B.S. degree in school of Communication and Information Engineering, Shanghai University (SHU), in 2015. He is currently pursuing the Ph.D. degree in signal and information processing from University of Science and Technology of China (USTC). His research interests include natural language understanding, natural text generation and deep learning.
\end{IEEEbiography}

\begin{IEEEbiography}[{\includegraphics[width=1in,height=1.25in,clip,keepaspectratio]{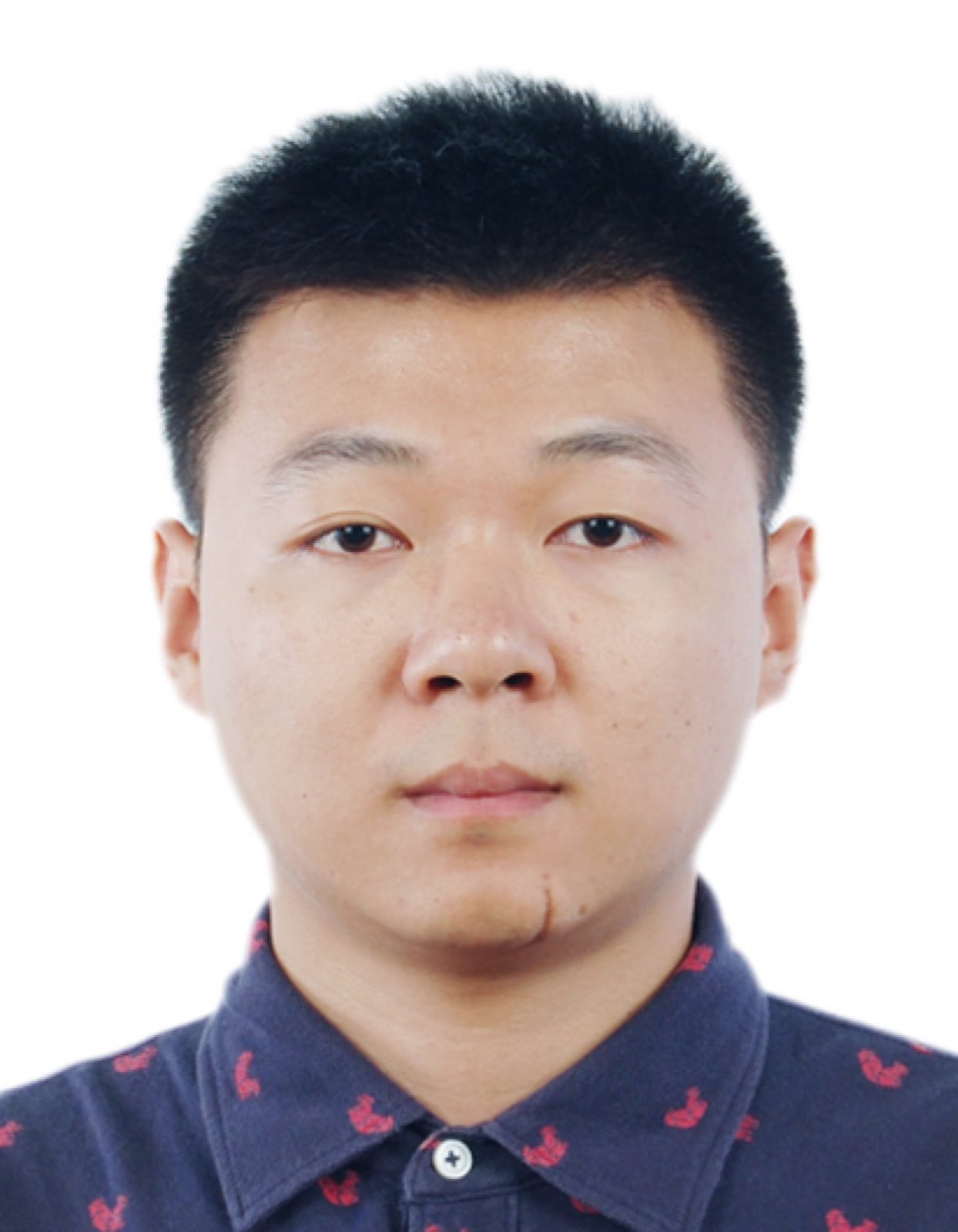}}]{Qian Chen} received the B.S. degree in electronic information engineering from University of Science and Technology of China (USTC), in 2013. He is currently pursuing the Ph.D. degree in signal and information processing from USTC. His research interests include natural language processing, speech synthesis and deep learning.
\end{IEEEbiography}

\begin{IEEEbiography}[{\includegraphics[width=1in,height=1.25in,clip,keepaspectratio]{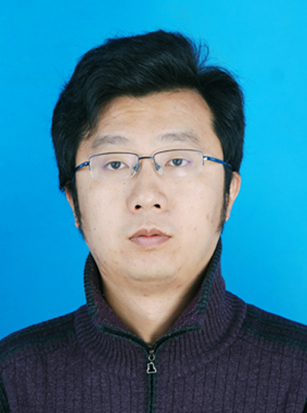}}]{Zhen-Hua Ling} (M¡¯10) received the B.E. degree
in electronic information engineering, the M.S. and
Ph.D. degree in signal and information processing
from the University of Science and Technology
of China, Hefei, China, in 2002, 2005, and 2008,
respectively. From October 2007 to March 2008,
he was a Marie Curie Fellow with the Centre for
Speech Technology Research (CSTR), University
of Edinburgh, Edinburgh, U.K. From July 2008
to February 2011, he was a joint Postdoctoral
Researcher with the University of Science and
Technology of China, Hefei, China, and iFLYTEK Co., Ltd., China. He is currently
an Associate Professor with the University of Science and Technology
of China. He also worked at the University of Washington, Seattle, WA, USA,
as a Visiting Scholar from August 2012 to August 2013. His research interests
include speech  processing, speech synthesis, voice conversion, and natural
language processing. He was the recipient of the IEEE Signal Processing Society
Young Author Best Paper Award in 2010. He is now an Associate Editor of IEEE/ACM
Transcations on Audio, Speech, and Language Processing.
\end{IEEEbiography}

%\begin{IEEEbiography}{Michael Shell}
%Biography text here.
%\end{IEEEbiography}

% if you will not have a photo at all:
%\begin{IEEEbiographynophoto}{John Doe}
%Biography text here.
%\end{IEEEbiographynophoto}

% insert where needed to balance the two columns on the last page with
% biographies
%\newpage

%\begin{IEEEbiographynophoto}{Jane Doe}
%Biography text here.
%\end{IEEEbiographynophoto}

% You can push biographies down or up by placing
% a \vfill before or after them. The appropriate
% use of \vfill depends on what kind of text is
% on the last page and whether or not the columns
% are being equalized.

%\vfill

% Can be used to pull up biographies so that the bottom of the last one
% is flush with the other column.
%\enlargethispage{-5in}

% that's all folks
\end{document}